\documentclass{article}

    \PassOptionsToPackage{numbers, compress}{natbib}

\usepackage[preprint]{arxiv_style}

\usepackage{tabularx} %
\usepackage{fontawesome5}
\usepackage{wrapfig}
\usepackage{todonotes}
\usepackage[utf8]{inputenc} %
\usepackage[T1]{fontenc}    %
\usepackage{hyperref}       %
\usepackage{url}            %
\usepackage{booktabs}       %
\usepackage{amsfonts}       %
\usepackage{nicefrac}       %
\usepackage{microtype}      %
\usepackage{xcolor}         %
\usepackage{graphicx}

\usepackage{xspace}
\usepackage[acronym]{glossaries-extra}
\setabbreviationstyle[acronym]{long-short}
\usepackage[most]{tcolorbox}
\usepackage{multirow}

\usepackage{algorithm}
\usepackage{algorithmic}

\usepackage{amsmath}
\usepackage[acronym]{glossaries-extra}
\setabbreviationstyle[acronym]{long-short}
\usepackage[most]{tcolorbox}
\usepackage{booktabs} %
\usepackage{tabularx} %
\usepackage{todonotes}
\usepackage{xspace}
\usepackage{amssymb}
\usepackage{amsfonts}
\usepackage{soul}
\usepackage{array}            %
\usepackage{subcaption}
\usepackage{caption}
\usepackage{tikz}

\definecolor{finalblue}{RGB}{0, 0, 0}
\definecolor{myblue}{RGB}{0, 0, 0}

\newcommand{\method}{\textsc{CLeGR}\xspace}
\newcommand{\tglm}{TEA-GLM\xspace}
\newcommand{\gret}{G-Retriever\xspace}

\newcommand{\grsagetok}{G-Token (GSAGE)\xspace}
\newcommand{\gattok}{G-Token (GAT)\xspace}
\newcommand{\softphithree}{Phi3-3.5B-SPT\xspace}

\newcommand{\softphifour}{Phi4-14B-SPT\xspace}

\newcommand{\softllama}{Llama3-8B-SPT\xspace}
\newcommand{\gcn}{GCN\xspace}
\newcommand{\gat}{GAT\xspace}
\newcommand{\sage}{GraphSAGE\xspace}
\newcommand{\llama}{Llama3-8B\xspace}
\newcommand{\phithree}{Phi3-3.5B\xspace}
\newcommand{\phifour}{Phi4-14B\xspace}

\tcbset{
  simplertakeaway/.style={
    colback=white,
    colframe=black,
    boxrule=0.4pt,
    arc=1pt,
    left=4pt,
    right=4pt,
    top=4pt,
    bottom=4pt,
    fonttitle=\bfseries,
  }
}

\newacronym{gnn}{GNN}{Graph Neural Network}
\newacronym{llm}{LLM}{Large Language Model}
\newacronym{glm}{GLM}{Graph Language Model}

\title{A Graph Talks, But Who's Listening?\\ Rethinking Evaluations for Graph-Language Models}
\workshoptitle{}

\author{%
  Soham~Petkar\textsuperscript{1,3}\footnotemark[1]
  \And
  Hari~Aakash~K\textsuperscript{1}\footnotemark[1]
  \And
  Anirudh~Vempati\textsuperscript{1}
  \AND
  Akshit~Sinha\textsuperscript{1}
  \And
  Ponnurangam~Kumaraguru\textsuperscript{1}
  \And
  Chirag~Agarwal\textsuperscript{2}
}

\begin{document}
\maketitle

\renewcommand{\thefootnote}{\fnsymbol{footnote}}
\footnotetext[1]{\textbf{Equal contribution.} \texttt{soham.petkar@plaksha.edu.in}, \texttt{hariaakash.k@research.iiit.ac.in}.}
\renewcommand{\thefootnote}{\arabic{footnote}}

\begingroup
\renewcommand\thefootnote{}\footnotetext{%
\textbf{Affiliations:} 
\textsuperscript{1} IIIT Hyderabad, India,\;
\textsuperscript{2} University of Virginia, USA,\;
\textsuperscript{3} Plaksha University, India}
\addtocounter{footnote}{-1}
\endgroup

\begin{center}
\raisebox{-1pt}{\faDatabase} \href{https://huggingface.co/datasets/tenseisoham/CLEGR}{\texttt{CLEGR Dataset}} \quad 
\raisebox{-1pt}{\faGithub} \href{https://github.com/rethinking-graph-language-evals/CLEGR}{\texttt{rethinking-graph-language-evals}}
\end{center}

\begin{abstract}
Developments in Graph-Language Models (GLMs) aim to integrate the structural reasoning capabilities of Graph Neural Networks (GNNs) with the semantic understanding of Large Language Models (LLMs). However, we demonstrate that current evaluation benchmarks for GLMs, which are primarily repurposed node-level classification datasets, are insufficient to assess multimodal reasoning. Our analysis reveals that strong performance on these benchmarks is achievable using unimodal information alone, suggesting that they do not necessitate graph-language integration. To address this evaluation gap, we introduce the \method (Compositional Language-Graph Reasoning) benchmark, designed to evaluate multimodal reasoning at various complexity levels. Our benchmark employs a synthetic graph generation pipeline paired with questions that require joint reasoning over structure and textual semantics. We perform a thorough evaluation of representative GLM architectures and find that soft-prompted LLM baselines perform on par with GLMs that incorporate a full GNN backbone. This result calls into question the architectural necessity of incorporating graph structure into LLMs. We further show that GLMs exhibit significant performance degradation in tasks that require structural reasoning. These findings highlight limitations in the graph reasoning capabilities of current GLMs and provide a foundation for advancing the community toward explicit multimodal reasoning involving graph structure and language.
\end{abstract}

\vskip -0.1in

\begin{figure*}[h]
  \centering
  \includegraphics[width=0.8\linewidth]{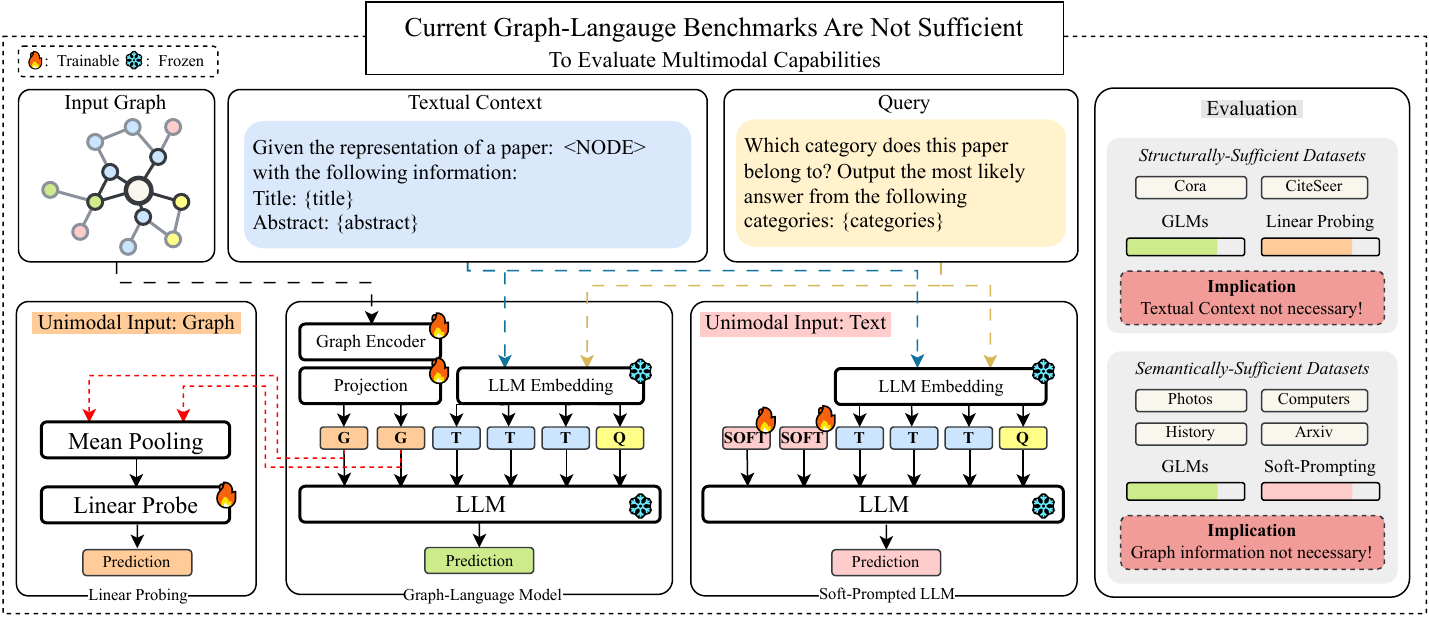}
  \captionsetup{font=footnotesize}
  \caption{Current graph-language benchmarks are insufficient for evaluating multimodal reasoning. We find these benchmarks can be solved using a single modality: linear probing on the graph tokens alone matches GLMs on structurally-sufficient datasets, while a soft-prompted LLM using text alone achieves the same on semantically-sufficient datasets. This shows models are not required to integrate both graph and text to succeed.}
  \label{fig:enter-label}
\end{figure*}

\section{Introduction}

The success of Vision-Language Models (VLMs) such as GPT-4V~\citep{yang2023dawnlmmspreliminaryexplorations} and LLaVa~\citep{liu2023visualinstructiontuning} demonstrates the transformative potential of integrating different data modalities to improve complex reasoning capabilities through visual question answering, image captioning, and multimodal instruction following~\citep{dai2023instructblipgeneralpurposevisionlanguagemodels}. Inspired by this paradigm, recently there has been an increasing interest in using LLMs for graph-based applications~\citep{li2024surveygraphmeetslarge}. Depending on the role of LLMs and their interaction with graph neural networks (GNNs), such techniques can be classified into treating LLMs as the final component for prediction (LLM as Predictor)~\citep{chen2024llagalargelanguagegraph, he2024gretrieverretrievalaugmentedgenerationtextual, he2024linkgptteachinglargelanguage, wang2024llmszeroshotgraphlearners}, treating LLMs as the feature extractor for GNNs (LLM as Encoder)~\citep{chien2022nodefeatureextractionselfsupervised, he2024harnessingexplanationsllmtolminterpreter, liu2024alltraininggraphmodel}, or aligning the latent space of LLMs with GNNs (LLM as Aligner)~\citep{ jin2023pattonlanguagemodelpretraining,zhao2023learninglargescaletextattributedgraphs}.

Similarly, according to the relationship between graph and text presented in the application, the scenarios can be categorized into pure graphs, Text-Paired graphs (entire graphs paired with a descriptive text summary)~\cite{jin2024largelanguagemodelsgraphs}, and Text-Attributed Graphs (TAGs). TAGs are prevalent in real-world applications where nodes represent textual entities like documents or papers, and edges capture relationships between them \cite{hu2021opengraphbenchmarkdatasets,yang2023graphformersgnnnestedtransformersrepresentation}. The combination of textual attributes with the graph structure has significantly improved representation learning for various applications ranging from recommendation systems to social networks~\citep{Zhu_2021}, with recent work further demonstrating that integrating textual semantics is critical for generalization in foundational models~\citep{arun2025semmasemanticawareknowledge}.

To effectively exploit the rich semantic information within TAGs, the LLM-as-predictor framework has emerged as a particularly promising direction~\citep{perozzi2024letgraphtalkingencoding}. Notably, LLM-as-predictor approaches have demonstrated zero-shot transfer capabilities across datasets, allowing models trained on a single dataset to generalize effectively to unseen graphs with different features and graph structure~\citep{wang2024llmszeroshotgraphlearners}.

In this work, we focus primarily on LLM-as-predictor approaches, as they represent a direct application of language models to graph question-answering tasks and align with our goal of assessing graph-language multimodal capabilities. Henceforth, we refer to this approach as \textit{Graph-Language Models (GLMs)} throughout the rest of the manuscript.

We first show that including node classification datasets in graph-language benchmarks does not provide a good proxy for tasks requiring a combination of structural and semantic knowledge. By training unimodal baselines which use either the graph structure, or the text attributes associated with the nodes, we achieve performance comparable to GLMs, questioning the utility of these datasets for assessing multimodal capabilities. To accurately assess these capabilities, we further introduce a new Graph
Question-Answering benchmark, \method, specifically designed so that both semantic and structural understanding are necessary to accurately answer the questions. Our contributions are summarized as follows:

1. We demonstrate that current GLMs can achieve strong performance on existing benchmarks by relying solely on graph or language modalities: On \textit{semantically-sufficient} datasets, we show that using a graph encoder as a backbone does not provide any advantage over using only text attributes for classification. On \textit{structurally-sufficient} datasets, we use linear probing~\citep{alain2018understandingintermediatelayersusing} to show that graph-encoder representations are sufficient to achieve performance comparable to the full GLM setup, questioning the utility of these datasets in evaluating the multimodal interplay of graphs and language.

2. We introduce \method (Compositional Graph-Language Reasoning), a synthetic benchmark explicitly constructed to assess multimodal graph-language reasoning. \method spans over 1,000 diverse graphs and 54,000 questions designed at multiple levels of structural reasoning and reasoning. We also show that while GLMs saturate performance on \method's retrieval tasks, their performance drops significantly on tasks requiring graph reasoning, and is on par with soft-prompted LLMs.

\section{Preliminaries}
\label{sec:prelims}

\newcommand{\concat}{{\big\Vert}}

\newcommand{\gnn}{{\text{GNN}}}
\newcommand{\pool}{{\text{Pool}}}

\newcommand{\llm}{{\text{LLM}}}
\newcommand{\wordemb}{{W}}
\newcommand{\vocab}{{L}}
\newcommand{\llmhdim}{{d}}

\newcommand{\glm}{{\text{GLM}}}
\newcommand{\proj}{{M_P}}
\newcommand{\predict}{\text{Predict}}

\newcommand{\textgraph}{{\mathcal{G}}}
\newcommand{\vertset}{{\mathcal{V}}}
\newcommand{\numvert}{{N}}
\newcommand{\edgeset}{{\mathcal{E}}}
\newcommand{\numedge}{{M}}
\newcommand{\featset}{{F}}
\newcommand{\vertfeatset}{{F^{\vertset}}}
\newcommand{\edgefeatset}{{F^{\edgeset}}}
\newcommand{\node}{{n}}
\newcommand{\feat}{{f}}
\newcommand{\vertfeat}{f^{\vertset}}
\newcommand{\edgefeat}{f^{\edgeset}}

\newcommand{\quesmark}{{q}}
\newcommand{\gt}{{g}}
\newcommand{\pred}{{\hat{y}}}
\newcommand{\textualize}{\text{Text}}

\subsection{Graph-Language Models}
Formally, a graph-language model $\glm=(M_l, M_g, \proj)$ comprises three main components: $M_l$ is an LLM, $M_g$ is the graph encoder, and $\proj$ is the linear projector that aligns graph and text representations. Let $\textgraph=(\vertset,\edgeset,\vertfeatset,\edgefeatset)$ be a Text Attributed Graph (TAG), where $\vertset$ is the set of nodes, $\edgeset$ is the set of edges between nodes, $\numvert=|\vertset|$ is the number of nodes in the graph, $\numedge=|\edgeset|$ is the number of edges, $\vertfeatset=\{\vertfeat_1,\vertfeat_2,\dots,\vertfeat_{\numvert}\}$ is the set of textual features of each node, and $\edgefeatset=\{\edgefeat_1,\edgefeat_2,\dots,\edgefeat_\numedge\}$ is the set of textual features of each edge. 

For node-level question-answering tasks, given a graph $\textgraph$, a node $\node_i \in\vertset$, its textual features $\feat_i\in \featset$, and a textual question $\quesmark$, the prediction made by the $\glm$ is:
\begin{align}
\hat{y} &= \text{Predict}_\glm(\textgraph,\node_i,\feat_i,\quesmark) \label{eq:node-level}\\
&= M_l\left(
     \proj(M_g(\textgraph,\node_i)) \concat 
     \wordemb(\feat_i) \concat
     \wordemb(\quesmark),
\right) \nonumber
\end{align}
where $\wordemb \in \mathbb{R}^{|\vocab|\times\llmhdim}$ is the word embedding matrix of the LLM ($\vocab$ denotes the token vocabulary and $\llmhdim$ the LLM's hidden dimension), and $\concat$ denotes concatenation of token sequences. Here, $\hat{y}, \quesmark, \feat_i \in \vocab^*$ where $\vocab^*$ represents sequences of tokens from the vocabulary. The ground truth $\gt$ also belongs to $\vocab^*$.

For graph-level question-answering tasks, given a graph $\textgraph$ and a textual question $\quesmark$, the GLM makes a prediction as:
\begin{align}
\hat{y} &= \text{Predict}_\glm(\textgraph,\quesmark) \label{eq:graph-level}\\
&= M_l\left(
     \proj(\text{Pool}(M_g(\textgraph))) \concat 
     \wordemb(\feat) \concat
     \wordemb(\quesmark)
\right) \nonumber
\end{align}

where $\text{Pool}(\cdot)$ aggregates node embeddings from $M_g(\textgraph,0), \ldots, M_g(\textgraph,\numvert-1)$ to produce a single graph-level representation, and $\feat$ represents the concatenated textual features of the entire graph.

\subsection{Soft Prompting}
To isolate the contribution of graph encoders in GLMs, we use soft prompting as a baseline. Soft Prompting \cite{lester2021powerscaleparameterefficientprompt} introduces learnable prompt tokens which are concatenated with the embeddings of the input given to the LLM. We note here that a GLM can be viewed as a soft-prompt where a graph encoder is learnt instead of token embeddings. A soft-prompted LLM consists of components $(M_l, \mathbf{s})$, where $M_l$ is the same LLM used in GLMs and $\mathbf{s} \in \mathbb{R}^{\llmhdim}$ is the trainable soft-prompt vector.
For a given task, the soft-prompted LLM prediction is:
\begin{align}
\hat{y}_{\text{soft}} &= M_l\left(
     \mathbf{s} \concat
     \wordemb(\feat_i) \concat
     \wordemb(\quesmark)
\right) \label{eq:soft-prompt}
\end{align}

The soft-prompt vector $\mathbf{s}$ is trained using the same objective and training procedure as the GLM, effectively learning to encode task-relevant information without access to graph structure. This baseline allows us to determine whether a sophisticated graph encoder is required to achieve performance gains or it can be achieved through simple parameter optimization in the language model space.

\section{Evaluating Current Benchmarks for Graph-Language Tasks}\label{sec:ncbench}

In this section, we primarily investigate the following questions: \textbf{RQ1:} Are node classification datasets a sufficient test for graph-language multimodality? \textbf{RQ2:} Do both GNN and LLM components of GLMs contribute to their strong performance on these datasets?

\subsection{Experimental Setup}
\label{sec:setup3}
We evaluate all models on six widely-used TAG datasets: Cora~\cite{mccallum2000automating}, CiteSeer~\cite{yang2016revisitingsemisupervisedlearninggraph}, Computers (Amazon Computers), Photo (Amazon Photos), History~\cite{shchur2019pitfallsgraphneuralnetwork}, and Arxiv~\cite{hu2021opengraphbenchmarkdatasets}, spanning diverse domains and graph structures. These datasets are specifically selected because they form the backbone of current graph-language evaluation practices, appearing across multiple prominent benchmarks including GLBench~\cite{li2024glbenchcomprehensivebenchmarkgraph}, GraphFM~\cite{xu2024graphfmcomprehensivebenchmarkgraph}, TAG~\cite{yan2023a}, and Planetoid~\cite{yang2016revisitingsemisupervisedlearninggraph}.

\noindent\textbf{Graph-Language Models.} We evaluate two prominent GLM architectures: (1) \tglm ~\cite{wang2024llmszeroshotgraphlearners}, which performs zero-shot graph learning by encoding graph structure through textual descriptions and leveraging LLMs for reasoning, and (2) GraphToken~\cite{perozzi2024letgraphtalkingencoding}, which learns discrete graph tokens to represent structural information and integrates them with language model processing. Both GLMs are equipped with GraphSAGE as the backbone, and each GLM is paired with both \llama and \phithree backbones to assess consistency across different language model scales.

\noindent\textbf{Graph-Only Baselines.} To isolate the contribution of structural information, we employ strong traditional GNNs that rely exclusively on graph topology: 
(1) GAT~\cite{veličković2018graphattentionnetworks}, which uses attention mechanisms to weigh the importance of neighboring nodes, 
(2) GCN~\cite{kipf2017semisupervisedclassificationgraphconvolutional}, which performs localized first-order approximations of spectral convolutions, and 
(3) GraphSAGE~\cite{hamilton2018inductiverepresentationlearninglarge}, which generates node embeddings by sampling and aggregating features from node neighborhoods.

\noindent\textbf{Language Only Baselines.} To isolate the contribution of textual information, we use soft-prompted LLMs~\cite{lester2021powerscaleparameterefficientprompt}. These models use identical \llama and \phithree backbones but operate only on the node text, augmented with trainable prompt vectors.

We employ identical training procedures across all models and report results across five random seeds. Additional implementation details are deferred to Appendix ~\ref{app:models}, sections ~\ref{app:glm_training_details}, ~\ref{app:gnn_training_details},~\ref{sec:node-eval}, ~\ref{app:compute}.

\begin{table}[t]
\centering
\caption{Accuracy on Node Classification tasks reveals two dataset categories: \textit{semantically-sufficient} datasets (Computers, Photo, History, Arxiv) where soft-prompted LLMs approximate GLM performance, indicating textual content alone suffices for classification; and \textit{structurally-sufficient} datasets (Cora, CiteSeer) where GNNs dominate and soft-prompted LLMs fail, suggesting graph structure is critical while semantic reasoning capabilities remain underutilized in current GLM evaluations.}
\vspace{0.1in}
\small
\resizebox{\linewidth}{!}{%
\begin{tabular}{l|cccc|cc}
\toprule
\textbf{Model} & \textbf{Computers} & \textbf{Photo} & \textbf{History} & \textbf{Arxiv} & \textbf{Cora} & \textbf{CiteSeer} \\
\midrule
\multicolumn{7}{c}{\itshape GNNs} \\
\gat              & \underline{$93.67_{\pm 0.28}$} & \underline{$96.51_{\pm 0.20}$} & $82.81_{\pm 0.74}$ & $73.30_{\pm 0.18}$ & $86.05_{\pm 1.37}$ & $71.12_{\pm 0.84}$ \\
\gcn              & $\mathbf{93.94_{\pm 0.13}}$ & $95.74_{\pm 0.10}$ & $82.91_{\pm 0.45}$ & $73.53_{\pm 0.12}$ & \underline{$86.98_{\pm 0.95}$} & \underline{$72.14_{\pm 0.67}$} \\
\sage             & $93.11_{\pm 0.23}$ & $\mathbf{96.54_{\pm 0.15}}$ & $83.24_{\pm 0.82}$ & $73.00_{\pm 0.28}$ & $\mathbf{87.31_{\pm 0.81}}$ & $\mathbf{72.26_{\pm 0.70}}$ \\
\midrule
\multicolumn{7}{c}{\itshape Graph--Language Models} \\
\tglm\,\llama      & $73.10_{\pm 1.07}$ & $70.51_{\pm 1.33}$ & $81.56_{\pm 5.39}$ & $73.08_{\pm 0.00}$ & $82.26_{\pm 1.23}$ & $48.05_{\pm 1.28}$ \\
\tglm\,\phithree    & $69.88_{\pm 0.61}$ & $64.81_{\pm 4.13}$ & $81.63_{\pm 0.58}$ & $67.38_{\pm 1.31}$ & $82.49_{\pm 1.17}$ & $42.08_{\pm 3.36}$ \\
\grsagetok\,\llama & $76.13_{\pm 0.58}$ & $76.61_{\pm 0.92}$ & $\mathbf{85.91_{\pm 0.27}}$ & \underline{$75.49_{\pm 0.28}$} & $86.72_{\pm 0.93}$ & $53.23_{\pm 0.53}$ \\
\grsagetok\,\phithree & $72.38_{\pm 0.90}$ & $73.50_{\pm 2.55}$ & \underline{$85.15_{\pm 0.31}$} & $71.38_{\pm 0.19}$ & $86.60_{\pm 1.12}$ & $44.69_{\pm 1.22}$ \\
\midrule
\multicolumn{7}{c}{\itshape Soft-Prompted LLMs} \\
\softllama        & $74.34_{\pm 0.63}$ & $74.90_{\pm 0.57}$ & \underline{$84.99_{\pm 0.66}$} & $\mathbf{76.03_{\pm 0.45}}$ & $28.69_{\pm 2.70}$ & $18.21_{\pm 0.26}$ \\
\softphithree     & $69.74_{\pm 0.26}$ & $70.71_{\pm 3.81}$ & $84.55_{\pm 0.43}$ & $72.04_{\pm 1.15}$ & $29.74_{\pm 1.12}$ & $18.28_{\pm 1.43}$ \\
\bottomrule
\end{tabular}
}

\label{tab:node_classification}
\end{table}

 \subsection{Analysis of Modality Contribution}
Informed by the performance of unimodal (graph-only vs language-only) baselines, we categorize the datasets into two groups: \textit{(a) semantically-sufficient}  and \textit{(b) structurally-sufficient}.

\looseness=-1 First, on the Computer, Photo, History, and Arxiv datasets, which we term \textit{semantically-sufficient}, we observe in Table~\ref{tab:node_classification} that soft-prompted LLMs achieve results that are highly competitive with GLMs across all datasets. GNNs outperform both GLMs and soft-prompted LLMs significantly on Computers and Photos, but are equally as good on History and Arxiv. These results suggest that, for these datasets, the semantic content present in the textual node attributes is sufficient to achieve performance equivalent to a multimodal model (incorporating the graph structure with the semantic content), making a graph encoder not strictly necessary for achieving high performance.

\begin{figure}[t]
    \centering
    \includegraphics[width=0.6\linewidth]{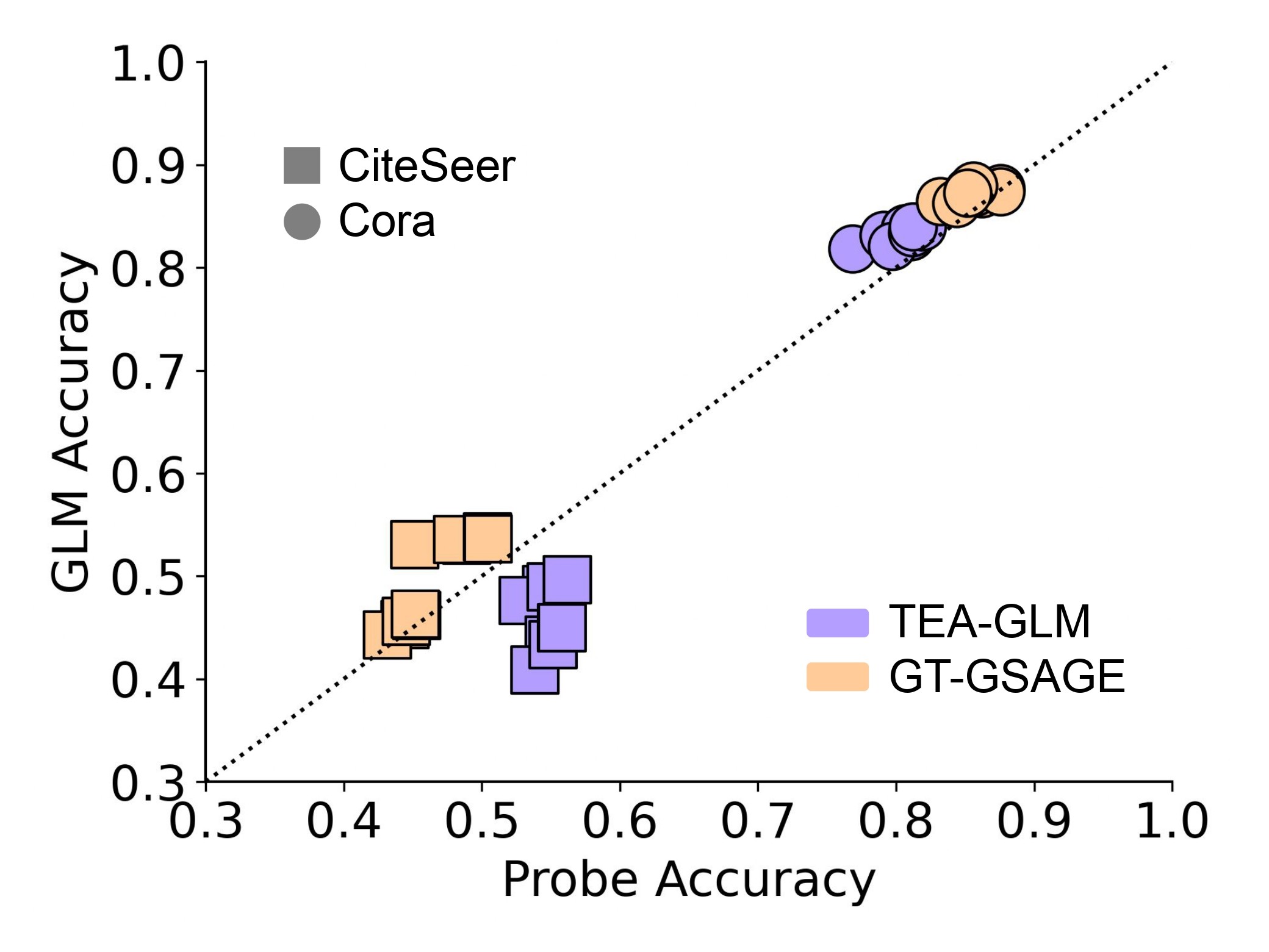}
    \caption{Linear probe accuracy closely matches full GLM performance on structurally-sufficient datasets, with Pearson correlation $r=0.9643$, showing that graph encoders capture all task-relevant information while LLMs act merely as expensive decoder heads.}
    \label{fig:probe_plot}
\end{figure}

Our results on Cora and CiteSeer follow a different trend. In these datasets, we find that structural information alone can saturate performance. We term these datasets to be \textit{structurally-sufficient}. Simply providing the textual attributes related to a node is not enough to classify it accurately, as is shown by the low accuracies achieved by soft-prompted LLMs. These results suggest that the addition of textual attributes does not provide any extra gains in accuracy on top of what is already learned by the graph structure. Since LLMs alone cannot achieve strong performance in these datasets, we hypothesize that the large accuracy gains that GLMs achieve over LLMs can be attributed to their graph encoder. To test this hypothesis, we perform a detailed probing analysis in the next section.

\subsection{Probing Graph Tokens}\label{sec:probing}

To isolate and quantify the contribution of the graph encoder on \textit{structurally-sufficient} datasets, we perform a linear probing analysis. First, we take a fully trained and frozen GLM and pass the graph data through its graph encoder to extract the final node representations (\textit{i.e.,} the graph tokens). Next, we pass these graph tokens through the language model. A simple linear classifier is then trained on top of these frozen representations to perform the node classification task. Mathematically, the prediction made by the probe can be represented as:
\begin{equation}
\pred = \arg\max_i \left(W \cdot \text{flatten}\left(\proj\big(M_g(\textgraph)\big)\right) + b\right)_i
\label{eq:probe}
\end{equation}
where $W \in \mathbb{R}^{c \times d'}$ and $b \in \mathbb{R}^{c}$ denote the classifier weights and bias, $c$ is the number of output classes, and $d'$ is the dimensionality of the flattened embedding representation. This minimal setup isolates the graph encoder and projection module to quantify linearly separable task-specific information without the influence of the LLM ($M_l$).

Our probing results (shown in Figure~\ref{fig:probe_plot}) on Cora and CiteSeer using \tglm and \grsagetok reveal that a simple linear classifier applied directly to the projector outputs achieves accuracy that nearly matches the full GLM performance. %
This finding provides strong evidence that on \textit{structurally-sufficient} datasets, the graph encoder captures all the critical information required for the task, suggesting that the LLM is functionally similar to a very large decoder network. Therefore, the textual-semantic reasoning capabilities of LLMs remain unutilized for these datasets.

\begin{tcolorbox}[simplertakeaway, title=Takeaway \#1]
Our analysis highlights a significant gap in existing benchmarks: a lack of datasets that are neither purely \textit{semantically-sufficient} nor \textit{structurally-sufficient}, but instead require the integration of both modalities to test true graph-language multimodality.
\end{tcolorbox}

\begin{figure*}[t]
  \centering
  \includegraphics[width=0.9\linewidth]{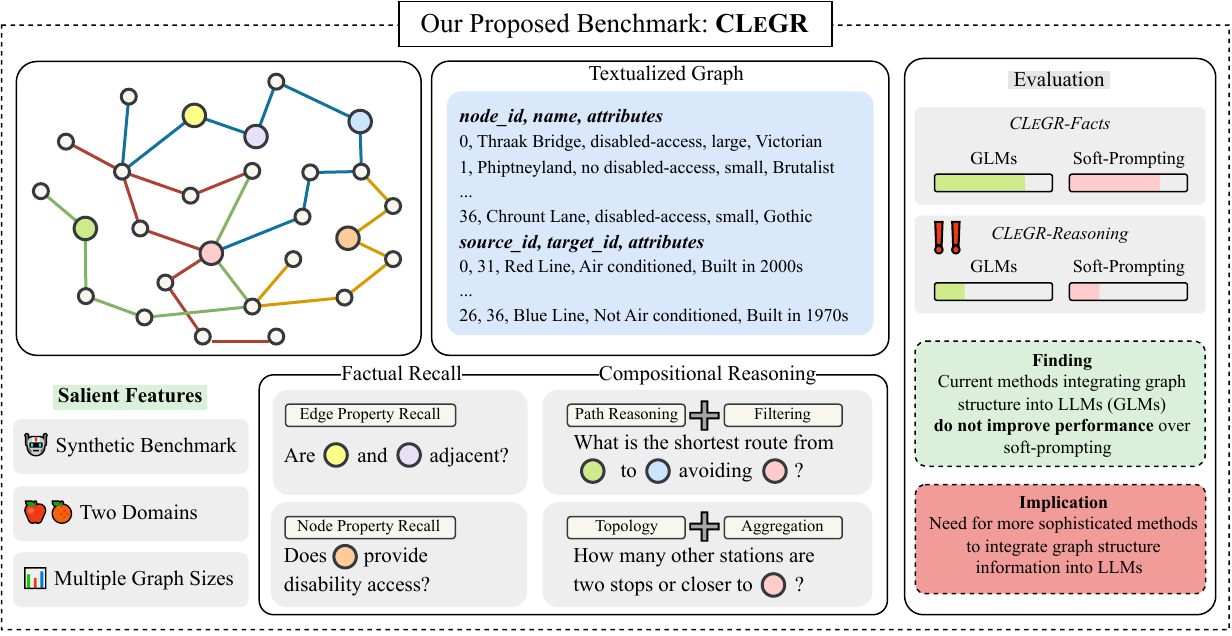}
\caption{Illustration of our evaluation framework and the \method benchmark. \method addresses limitations of existing benchmarks using synthetic graphs with explicit node and edge attributes, covering tasks from factual recall (\method-Facts) to compositional reasoning (\method-Reasoning) across filtering, aggregation, path reasoning, and topology. Evaluation shows that GLMs match baselines on fact-based tasks but fail to outperform soft-prompted LLMs on reasoning, indicating insufficient graph-language integration.}

  \label{fig:clegr}
\end{figure*}

\section{\method: Compositional Language-Graph Reasoning}

Having identified limitations in the current evaluation setup for graph-language models, we now introduce a framework that is neither \textit{semantically-sufficient} nor \textit{structurally-sufficient}. To accurately assess multimodal capabilities, we construct \textbf{\method}, a benchmark suite to enable reasoning over graph structures with explicit node and edge textual attributes.

\subsection{Overview of \method}

\looseness=-1\method's design is guided by three core principles to preclude unimodal solutions: i) \textit{Structural Dependency}: Tasks require multi-hop reasoning over graph topology that cannot be inferred from the node text alone, making vanilla language models insufficient as they lack access to structural relationships beyond immediate textual context~\cite{yasunaga-etal-2021-qa};
ii) \textit{Semantic grounding:} Questions that require natural language understanding capabilities, making traditional GNNs inadequate as they operate on numerical representations without semantic comprehension~\cite{hamilton2018representationlearninggraphsmethods,ying2019gnnexplainergeneratingexplanationsgraph}; and
iii) \textit{Compositional complexity:} Tasks combine multiple reasoning steps that blend property lookup with logical inference, creating challenges that benefit from integrating both structural and semantic information throughout the reasoning~\cite{yang2018hotpotqadatasetdiverseexplainable}.

\method is constructed as an extension of the foundational $\textsc{CLeVR}$-Graph dataset~\cite{mack2018clevrgraph}, which introduced synthetic graph question answering tasks using fictional subway system graphs. The synthetic design of stations, lines, and connections helps to eliminate pre-training confounds, ensuring language models cannot rely on memorized knowledge.

\subsection{\method Design and Structure}

The \method benchmark (see Figure~\ref{fig:clegr}) contains 1000 synthetic subway graphs split into two subsets: \method-Facts (500 graphs) and \method-Reasoning (500 graphs), each using a 3:1:1 train/validation/test split.

\looseness=-1\method-Facts is a purely retrieval-based benchmark, 
requiring node/edge property lookup without graph structure reasoning (For example, \textit{What music is played on Station X?}). This subset generates exactly 22,000 questions (44 per graph from 22 templates) and serves as a baseline to assess LLM performance disparity when reasoning over graph structure is not required.

\method-Reasoning extends beyond current TAG benchmarks by adding compositional reasoning over graph structure. It contains 32,248 questions generated from 34 templates, with natural filtering removing structurally invalid instances. Questions demand multi-step inference and graph traversal (e.g., \textit{What is the shortest path between Station A and B using only air-conditioned lines?}).
\method-Reasoning systematically covers four reasoning types (Filtering, Aggregation, Path Reasoning, Topology) across three scopes (node, edge, subgraph-level). The compositional design combines multiple reasoning types to create complex multi-step problems that probe different facets of graph-language reasoning capabilities.
Detailed structural statistics, template examples, and dataset construction process are present in Appendix~\ref{app:clegr_construction} and Appendix~\ref{sec:appendix_dataset}.

\section{Evaluating GLMs on \method}\label{clegreval}
Equipped with a benchmark that necessitates the incorporation of graph structure and text semantics, we now evaluate GLMs on \method. Specifically, we answer the following questions: \textbf{RQ3:} Does incorporating structural information into LLMs provide performance gains over soft-prompting LLMs on tasks requiring multimodal reasoning? \textbf{RQ4:} Do GLMs provide better zero-shot generalization to other domains? \textbf{RQ5:} How does GLM performance scale with increasing graph size?

\subsection{Experimental Setup}
\subsubsection{Models} We evaluate several model categories to comprehensively assess graph reasoning capabilities. First, we include the GLMs introduced in Section~\ref{sec:setup3}. Second, we add GLMs with retrieval-augmented approaches, specifically \gret~\cite{he2024gretrieverretrievalaugmentedgenerationtextual}, which enhances task performance through subgraph and textual retrieval to extract relevant nodes and edges for answering questions. Following \gret's setup, we adapt all GLMs to our tasks by including graphs in textualized format, ensuring consistent input prompts across models.  
To evaluate GLM performance across different backbones, we also evaluate GraphToken with a GAT backbone.
We also include LLM-only baselines, adding \phifour to our initial set of LLMs from Section 3.1 to analyze scaling effects as model size increases.

\subsubsection{Training} All experiments use identical hardware configurations with consistent batch sizes and learning rates within model categories. We employ greedy decoding for generation and average results across 5 random seeds for statistical reliability. 
Importantly, for \method, we employ the pooling operation from Equation~\ref{eq:graph-level} to pass the complete graph representation to all models, as reasoning questions may target any combination of nodes, edges, and subgraphs, requiring full structural context. Additional training details are provided in the Appendix~\ref{app:evaluation_details}.

\subsubsection{Evaluation Metrics.} Performance is evaluated using overall accuracy across all question types. A prediction is considered correct only when it precisely matches the ground truth. We provide detailed performance breakdowns across our two major subsets (Facts vs. Reasoning). Exact evaluation details are deferred to the Appendix.

\subsection{Results}\label{sec:results}

\begin{figure}
    \centering
    \includegraphics[width=\linewidth]{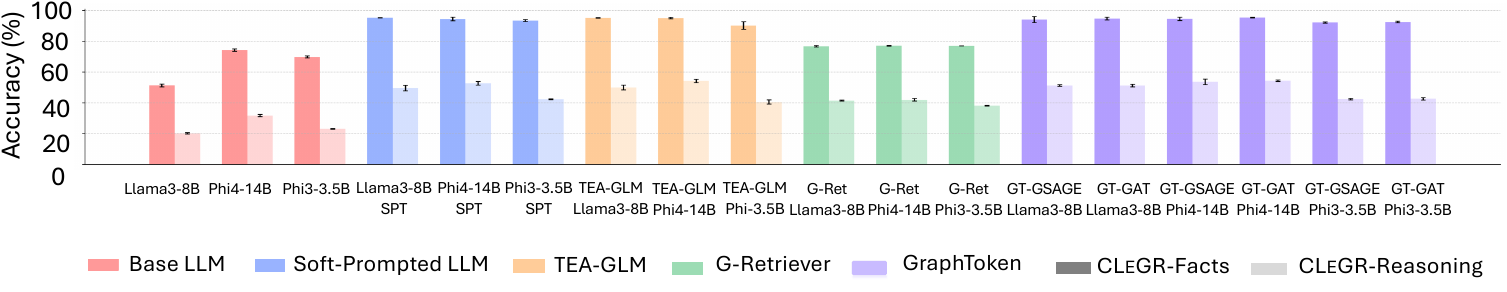}
    \caption{\method Results: GLMs achieve saturation on fact-based retrieval tasks but fail to outperform soft-prompted baselines on reasoning tasks requiring structural understanding, revealing a reliance on surface-level language patterns over structural graph understanding.}
    \label{fig:clegr_main}
\end{figure}

\subsubsection{RQ3: Does incorporating structural information into LLMs provide performance gains over soft-prompting LLMs on tasks requiring multimodal reasoning?} Our results, presented in Figure~\ref{fig:clegr_main}, show that despite using diverse architectural strategies for encoding structural information into LLMs, GraphToken and \tglm provide negligible performance gains, if any, over purely language-based soft-prompted baselines. Surprisingly, \gret, which uses a sophisticated subgraph retrieval mechanism intended to selectively provide relevant graph substructures to the model, suffers from a degradation rather than an improvement in performance. We hypothesize that \gret's performance degradation stems from potentially retrieving incorrect subgraphs, and not having sufficient context to answer questions correctly. These findings collectively indicate that current GLMs do not utilize multimodal inputs effectively; instead, they revert to powerful but ultimately unimodal textual processing, failing to integrate the structural data in a meaningful way.

\begin{figure}[t]
    \centering
    \begin{subfigure}{0.48\linewidth}
        \centering
        \includegraphics[width=\linewidth]{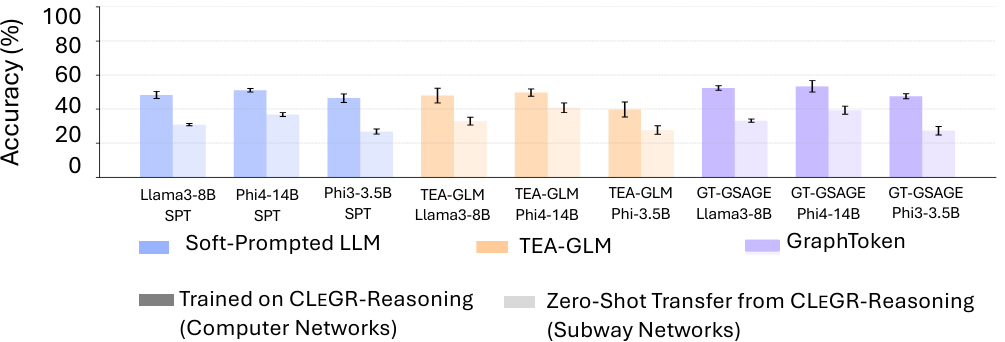}
        \caption{Zero-shot generalization from subway to computer network domains shows GLMs provide no transfer benefits compared to soft-prompted approaches, indicating structural encoders do not enhance cross-domain reasoning capabilities.}
        \label{fig:clegr_transfer}
    \end{subfigure}
    \hfill
    \begin{subfigure}{0.48\linewidth}
        \centering
        \includegraphics[width=\linewidth]{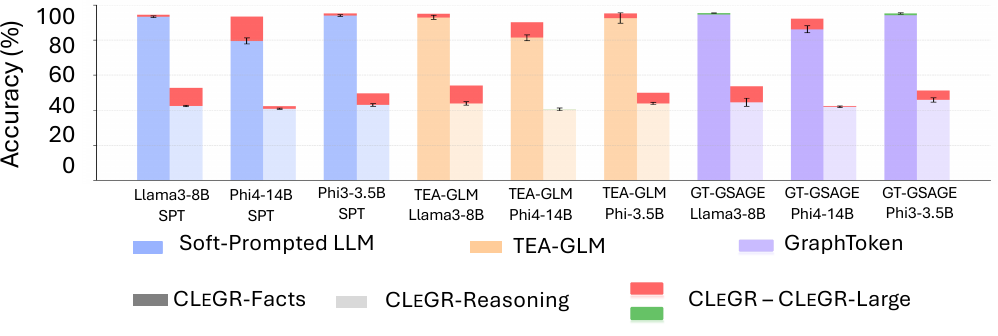}
        \caption{Results on graphs larger than standard \method demonstrate that increased structural complexity provides no advantage to GLMs over soft-prompted baselines, with both approaches showing comparable performance degradation.}
        \label{fig:clegr_large}
    \end{subfigure}
    \caption{Evaluation Results: (a) zero-shot transfer to a new semantic domain (\method-Computer-Networks), and (b) the impact of increasing graph size (\method-Large).}
    \label{fig:clegr_large_transfer}
\end{figure}

\subsubsection{RQ4: Do GLMs provide better zero-shot generalization to other domains?} To assess if the structural encoding in GLMs leads to better zero-shot performance, we evaluate their transfer capability from the original subway domain presented in \method-Reasoning to a new, structurally analogous Computer-Networks domain. The results are presented in Figure~\ref{fig:clegr_transfer}. Neither TEA-GLM nor GraphToken achieve significant performance improvements compared to soft-prompted LLMs, showcasing that the injection of structural information by GLMs does not have a significant impact on zero-shot transfer (More details in Appendix~\ref{app:computer_networks}).

\subsubsection{RQ5: How does GLM performance scale with increasing graph size?}
We test whether increasing the size of graphs present in \method show a scenario where GLMs outperform language-only baselines. For evaluation purposes, we introduce \method-Large, a modification of \method that contains graphs approximately three times larger (additional details presented in Appendix~\ref{app:graph_stats} Table~\ref{tab:graph_stats}). As graph complexity increased, both approaches exhibited nearly identical performance degradation on reasoning and factual tasks (see Figure~\ref{fig:clegr_large}). This parallel decline demonstrates that GLMs offer no inherent advantage for handling large graph structures, as their performance scales just like simpler, text-based methods. This result corroborates our findings on \textbf{RQ3} and \textbf{RQ4}, highlighting negligible advantages produced by GLMs.

\begin{tcolorbox}[simplertakeaway, title=Takeaway \#2]
GLMs fail to leverage their structural encoders for graph reasoning tasks. GLMs offer no significant advantage over soft-prompted baselines, and this parity holds even as graph size increases, or while performing zero-shot transfer. Our findings suggest GLMs' capabilites are primarily driven by the LLM's textual processing capabilities rather than an interplay of graph and text modalities.
\end{tcolorbox}

\section{Analyzing Representation Alignment}\label{subsec:cka}

\begin{figure}
    \centering
    \includegraphics[width=0.72\linewidth]{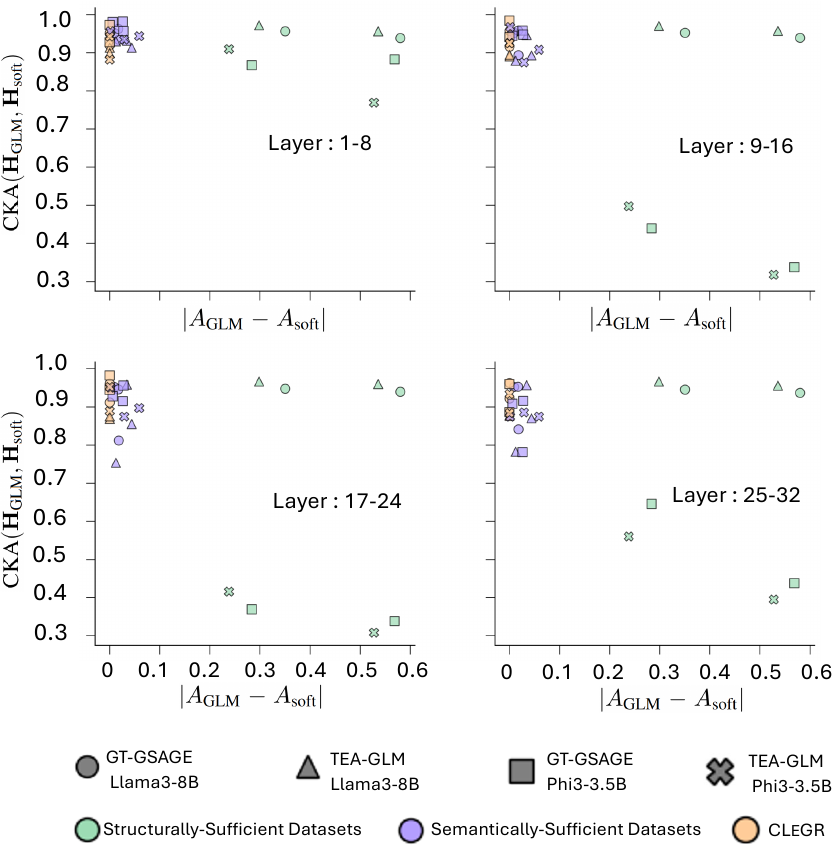}
    \caption{CKA aligns strongly when performance is similar. \textit{Semantically-sufficient} datasets show near-identical representations; \textit{Structurally-sufficient} datasets diverge in mid layers, aligning with soft-prompt failures. Each point denotes a (dataset, model, layer-group) triplet.}
    \label{fig:cka_scatter}
\end{figure}

Our findings in Section~\ref{sec:ncbench} and Section~\ref{clegreval} empirically demonstrate that simply soft-prompting LLMs matches or exceeds GLMs across all our evaluations (Table~\ref{tab:node_classification}, Figure~\ref{fig:clegr_main}, Figure~\ref{fig:clegr_transfer}, and Figure~\ref{fig:clegr_large}). In this section, we experimentally investigate if the cause behind similar performance metrics for different models is similar internal representations. To verify this, we employ Centered Kernel Alignment (CKA)~\cite{kornblith2019similarityneuralnetworkrepresentations} to measure representational overlap between GLMs and soft-prompts.  As shown in Figure~\ref{fig:cka_scatter}, we analyze the relationship between the performance gap $|A_{\text{GLM}} - A_{\text{soft}}|$ (x-axis), where where $A$ denotes accuracy and representational similarity $\text{CKA}(\mathbf{H}_{\text{GLM}},  \mathbf{H}_{\text{soft}})$ (y-axis), where $\mathbf{H}$ represents the activation over graph and soft-prompt tokens respectively.

We observe that most datasets cluster in the upper-left region of Figure~\ref{fig:cka_scatter} (low performance gap, high CKA). \textit{Semantically-sufficient} datasets and \method tasks maintain high CKA across all layers. \textit{Structurally-sufficient} datasets show lower CKA in middle layers (9-24), correlating with our observed low performance by soft-prompted baselines on these datasets. Our analysis highlights that GLMs learn different representations only when datasets are \textit{structurally-sufficient}, i.e., when the LLM's textual reasoning capabilities are underutilized (Section~\ref{sec:probing}).

\section{Discussion}

In this work, we first demonstrate that on current graph-language benchmarks that incorporate Node Classification datasets to evaluate multimodality, performance equivalent to GLMs can be achieved by using unimodal baselines, highlighting the need for new evaluation paradigms. To fill this gap, we introduce \method, a synthetic graph-language reasoning benchmark explicitly designed to assess multimodal integration. Evaluations on \method highlight limitations of current GLMs, showcasing that they provide negligible gains over soft-prompting LLMs, emphasizing the need for more sophisticated methods to integrate graph information into LLMs. While \method introduces a framework to evaluate multimodal capabilities, several limitations and avenues for future work remain. As GLMs continue to develop, their capacity to represent graphs are likely to increase, and our current evaluation may not encompass all their capabilities. We aim to continually refine and update \method to more effectively assess the graph-language understanding and reasoning abilities of emerging GLMs.

\bibliographystyle{plainnat}
\bibliography{arxiv_reference}

\appendix
 
\appendix

\section{Models}
\label{app:models}

\subsection{\tglm}
The TEA-GLM (Token Embedding-Aligned Graph Language Model)   ~\cite{wang2024llmszeroshotgraphlearners} methodology is a novel framework designed to enhance zero-shot graph machine learning by integrating Graph Neural Networks (GNNs) with instruction-fine-tuned Large Language Models (LLMs). It consists of two main stages: first, pretraining a GNN using enhanced self-supervised learning with feature-wise contrastive learning to align its node representations with LLM token embeddings, enabling the GNN to leverage the LLM’s pretrained knowledge; second, training a linear projector to transform these GNN representations into a fixed number of graph token embeddings, which are incorporated into a unified instruction for various graph tasks, without tuning the LLM.

\subsection{\gret}
G-Retriever~\cite{he2024gretrieverretrievalaugmentedgenerationtextual} is a framework for question answering on textual graphs, integrating Retrieval-Augmented Generation (RAG) with Graph Neural Networks (GNNs) and Large Language Models (LLMs) to enable users to "chat with their graph." It addresses complex queries on real-world textual graphs by first developing a Graph Question Answering (GraphQA) benchmark with diverse datasets like ExplaGraphs, Scenegraphs, and WebQSP. G-Retriever employs a novel RAG approach, formulating subgraph retrieval as a Prize-Collecting Steiner Tree optimization problem to efficiently select relevant graph parts, mitigating scalability issues and LLM hallucination. The retrieved subgraph is textualized and combined with the query, then processed by a frozen LLM with soft prompting for fine-tuned, contextually accurate responses across applications like scene graph understanding, common sense reasoning, and knowledge graph reasoning.

\subsection{Graph-Token (G-Token)}
In the Graph-Token methodology~\cite{perozzi2024letgraphtalkingencoding} \grsagetok employs GraphSAGE as the GNN encoder to process the graph’s structure, generating representations through neighborhood aggregation, while \gattok uses GAT (Graph Attention Network) as the GNN encoder, leveraging attention mechanisms to weigh node connections. The representations are then mapped by a trained linear projector into token embeddings. These tokens are prepended to the prompt of a frozen LLM.

\subsection{GNN Architectures}

Let a graph be $G = (\mathcal{V}, \mathcal{E}, X, Y)$, where $\mathcal{V}$ denotes the set of nodes, $\mathcal{E} \subseteq \mathcal{V} \times \mathcal{V}$ represents the set of edges, $X \in \mathbb{R}^{|\mathcal{V}| \times d}$ is the node feature matrix, with $|\mathcal{V}|$ representing the number of nodes and $d$ the dimension of the node features, and $Y \in \mathbb{R}^{|\mathcal{V}| \times C}$ is the one-hot encoded label matrix, with $C$ being the number of classes. Let $A \in \mathbb{R}^{|\mathcal{V}| \times |\mathcal{V}|}$ denote the adjacency matrix of $G$.

\subsubsection{Graph Convolutional Networks (GCN)}
GCNs ~\cite{kipf2017semisupervisedclassificationgraphconvolutional} update node embeddings using a normalized sum over neighboring features:
\begin{equation}
    h_v^{(l)} = \sigma \left( \sum_{u \in \mathcal{N}(v) \cup \{v\}} \frac{1}{\sqrt{\hat{d}_v \hat{d}_u}} h_u^{(l-1)} W^{(l)} \right),
\end{equation}
where $\hat{d}_v$ is the degree of node $v$ (including self-loops), $W^{(l)}$ is the trainable weight matrix at layer $l$, and $\sigma(\cdot)$ is an activation function such as ReLU.

\subsubsection{GraphSAGE}
GraphSAGE~\cite{hamilton2018inductiverepresentationlearninglarge} aggregates neighborhood information using a fixed aggregation function (e.g., mean):
\begin{equation}
    h_v^{(l)} = \sigma \left( h_v^{(l-1)} W_1^{(l)} + \left( \text{mean}_{u \in \mathcal{N}(v)} h_u^{(l-1)} \right) W_2^{(l)} \right),
\end{equation}
where $W_1^{(l)}$ and $W_2^{(l)}$ are trainable matrices and the neighbor embeddings are averaged.

\subsubsection{Graph Attention Networks (GAT)}
GATs~\cite{veličković2018graphattentionnetworks} apply masked self-attention over neighbors. The attention coefficient between nodes $v$ and $u$ is:
\begin{equation}
    \\ \frac{
        \exp\left(\text{LeakyReLU}\left( a^{\top} [W h_v^{(l-1)} \, \Vert \, W h_u^{(l-1)}] \right) \right)
    }{
        \sum_{r \in \mathcal{N}(v)} \exp\left(\text{LeakyReLU}\left( a^{\top} [W h_v^{(l-1)} \, \Vert \, W h_r^{(l-1)}] \right) \right)
    },
\end{equation}
and the node update is:
\begin{equation}
    h_v^{(l)} = \sigma \left( \sum_{u \in \mathcal{N}(v)} \alpha_{vu}^{(l)} W h_u^{(l-1)} \right),
\end{equation}
where $W$ is the shared weight matrix, $a$ is a learnable attention vector, and $\Vert$ denotes concatenation and  $\alpha_{vu}^{(l)}$ is the attention coefficient between nodes $v$ and $u$.

\subsection{Linear Probing}
\label{app:linear_probing}

Mathematically, the prediction made by the probe can be represented as:
\begin{equation}
\pred = \arg\max_i \left(W \cdot \text{flatten}\left(\proj\big(M_g(\textgraph)\big)\right) + b\right)_i
\label{eq:probe}
\end{equation}
where $W \in \mathbb{R}^{c \times d'}$ and $b \in \mathbb{R}^{c}$ denote the classifier weights and bias, $c$ is the number of output classes, and $d'$ is the dimensionality of the flattened embedding representation. This minimal setup isolates the graph encoder and projection module to quantify linearly separable task-specific information without the influence of the LLM ($M_l$).

\subsection{Training Details for GLMs and Soft-Prompt Baselines}
\label{app:glm_training_details}

All experiments are conducted using five random seeds: \texttt{0}, \texttt{42}, \texttt{1918}, \texttt{2004}, and \texttt{2024}. We use a fixed number of \texttt{10} graph tokens (projector output tokens) across all GLMs and soft-prompt baselines. The input format to the LLM follows the structure:
\begin{center}
\texttt{[\textless{}BOS\textgreater{} + 10x\textless{}G\textgreater{} + Context + Question + \textless{}EOS\textgreater{}]}
\end{center}
where \texttt{\textless{}G\textgreater{}} represents either the trainable graph tokens (in GLMs) or the fixed soft-prompt tokens (in soft-prompt baselines).

For both the \method and node classification tasks, all Graph-Language Models (GLMs) and soft-prompt variants are trained for a single epoch using the AdamW optimizer with a constant learning rate of \texttt{0.001} and a batch size of \texttt{1} during both training and evaluation.

All GLMs are implemented using a GraphSAGE~\cite{hamilton2018inductiverepresentationlearninglarge} backbone, with the exception of \gattok, which uses a GAT-based encoder. Furthermore, the \tglm models are pre-trained using 1000 PCA-projected features extracted from each LLM, following the protocol described in~\cite{wang2024llmszeroshotgraphlearners}.   (Table~\ref{tab:g-retriever-hparams} \& Table~\ref{tab:gat-hparams})

\subsection{GNN Training}
\label{app:gnn_training_details}

We follow the works of~\cite{luo2024classicgnnsstrongbaselines} for GNN baselines; we follow the same training setup, adopting their recommended hyperparameters and optimization configurations for each dataset. (Table~\ref{tab:gnn-hyperparams})

\subsection{Node Classification Evaluation}
\label{sec:node-eval}

For the \textbf{GNN baselines}, we adopt standard evaluation pipelines as used in prior work, including accuracy-based evaluation over ground-truth node labels. Specifically, following~\cite{luo2024classicgnnsstrongbaselines}, models are trained and evaluated on fixed data splits, and performance is reported as the mean and standard deviation across five random seeds.

For the \textbf{Graph-Language Models (GLMs)} and \textbf{soft-prompt models}, we adopt a flexible string-matching protocol to map generated textual responses to discrete class labels. Each dataset-specific evaluator contains a fixed list of candidate class names and defines a `match\_prediction' function that attempts to align the raw LLM output with one of the ground-truth labels. A prediction is considered correct if it begins with the correct class name (e.g., \texttt{"Asia"} matches \texttt{"Asia in the 20th century"}). Unmatched predictions are mapped to a special \texttt{None} class index. 

This string-based matching enables compatibility between natural language generation from LLMs and traditional classification metrics like accuracy. Final accuracy scores are computed by comparing matched predictions to the ground-truth class label.

\begin{table*}[t]
\centering
\caption{Hyperparameter configurations for GCN, GAT, and GraphSAGE across five benchmark datasets. `Norm' abbreviates normalization (LN: LayerNorm, BN: BatchNorm), and `LR' is the learning rate.}
\footnotesize
\renewcommand{\arraystretch}{1.15}
\setlength{\tabcolsep}{5.5pt}
\begin{tabular}{@{\extracolsep{\fill}}lcccccccc}
\toprule
\textbf{Model} & \textbf{Dataset} & \textbf{ResNet} & \textbf{Norm} & \textbf{Dropout} & \textbf{\#Layers} & \textbf{Hidden Dim} & \textbf{LR} & \textbf{Epochs} \\
\midrule
\multirow{5}{*}{GCN} 
& Cora      & False & None & 0.7 & 3 & 512 & 0.001 & 500 \\
& Citeseer  & False & None & 0.5 & 2 & 512 & 0.001 & 500 \\
& Computer  & False & LN   & 0.5 & 3 & 512 & 0.001 & 1000 \\
& Photo     & True  & LN   & 0.5 & 6 & 256 & 0.001 & 1000 \\
& History   & True  & LN   & 0.5 & 6 & 256 & 0.001 & 1000 \\
& Arxiv     & True  & BN   & 0.5 & 5 & 512 & 0.0005 & 2000 \\
\midrule
\multirow{5}{*}{GAT} 
& Cora      & True  & None & 0.2 & 3 & 512 & 0.001 & 500 \\
& Citeseer  & True  & None & 0.5 & 3 & 256 & 0.001 & 500 \\
& Computer  & False & LN   & 0.5 & 2 & 64  & 0.001 & 1000 \\
& Photo     & True  & LN   & 0.5 & 3 & 64  & 0.001 & 1000 \\
& History   & True  & LN   & 0.5 & 3 & 64  & 0.001 & 1000 \\
& Arxiv     & True  & BN   & 0.5 & 5 & 256 & 0.0005 & 2000 \\
\midrule
\multirow{5}{*}{GraphSAGE} 
& Cora      & False & None & 0.7 & 3 & 256 & 0.001 & 500 \\
& Citeseer  & False & None & 0.2 & 3 & 512 & 0.001 & 500 \\
& Computer  & False & LN   & 0.3 & 4 & 64  & 0.001 & 1000 \\
& Photo     & True  & LN   & 0.2 & 6 & 64  & 0.001 & 1000 \\
& History     & True  & LN & 0.2 & 6 & 64  & 0.001 & 1000 \\
& Arxiv     & True  & BN   & 0.5 & 4 & 256 & 0.0005 & 2000 \\
\bottomrule
\end{tabular}

\label{tab:gnn-hyperparams}
\end{table*}

\begin{table}[t]
\caption{GLM hyperparameter settings for using the \gret\ \& \grsagetok\ models. All configurations use consistent hidden/project dimensions and dropout.}
\label{tab:g-retriever-hparams}
\centering
\small
\renewcommand{\arraystretch}{1.05}
\setlength{\tabcolsep}{3pt} %
\begin{tabular}{@{} l l l c c c c c c @{}}
\toprule
\textbf{Dataset} & \textbf{GNN backbone} & \textbf{Task} & \textbf{In Dim} & \textbf{Hidden Dim} & \textbf{Out Dim} & \textbf{Proj Dim} & \textbf{Layers} & \textbf{Dropout} \\
\midrule
Cora              & GraphSAGE & Node  & 500 & 1024 & 1024 & 1024 & 3 & 0.5 \\
Citeseer          & GraphSAGE & Node  & 500 & 1024 & 1024 & 1024 & 3 & 0.5 \\
Arxiv             & GraphSAGE & Node  & 128 & 1024 & 1024 & 1024 & 3 & 0.5 \\
Computers         & GraphSAGE & Node  & 768 & 1024 & 1024 & 1024 & 3 & 0.5 \\
History           & GraphSAGE & Node  & 768 & 1024 & 1024 & 1024 & 3 & 0.5 \\
Photo             & GraphSAGE & Node  & 768 & 1024 & 1024 & 1024 & 3 & 0.5 \\
\method-Facts     & GraphSAGE & Graph & 768 & 1024 & 1024 & 1024 & 3 & 0.5 \\
\method-Reasoning & GraphSAGE & Graph & 768 & 1024 & 1024 & 1024 & 3 & 0.5 \\
\bottomrule
\end{tabular}
\end{table}

\begin{table}[t]
\caption{GLM hyperparameter settings for using the \gat\ models. All configurations use consistent hidden/project dimensions and dropout.}
\label{tab:gat-hparams}
\centering
\small
\renewcommand{\arraystretch}{1.05}
\setlength{\tabcolsep}{3pt} %
\resizebox{\linewidth}{!}{%
\begin{tabular}{l l l c c c c c c}
\toprule
\textbf{Dataset} & \textbf{GNN backbone} & \textbf{Task} & \textbf{In Dim} & \textbf{Hidden Dim} & \textbf{Out Dim} & \textbf{Proj Dim} & \textbf{Layers} & \textbf{Dropout} \\
\midrule
Cora              & GAT & Node  & 500 & 1024 & 1024 & 1024 & 3 & 0.5 \\
Citeseer          & GAT & Node  & 500 & 1024 & 1024 & 1024 & 3 & 0.5 \\
Arxiv             & GAT & Node  & 128 & 1024 & 1024 & 1024 & 3 & 0.5 \\
Computers         & GAT & Node  & 768 & 1024 & 1024 & 1024 & 3 & 0.5 \\
History           & GAT & Node  & 768 & 1024 & 1024 & 1024 & 3 & 0.5 \\
Photo             & GAT & Node  & 768 & 1024 & 1024 & 1024 & 3 & 0.5 \\
\method-Facts     & GAT & Graph & 768 & 1024 & 1024 & 1024 & 3 & 0.5 \\
\method-Reasoning & GAT & Graph & 768 & 1024 & 1024 & 1024 & 3 & 0.5 \\
\bottomrule
\end{tabular}
}
\end{table}

\subsection{Computing Infrastructure}
\label{app:compute}

All experiments are conducted on machines equipped with NVIDIA A100 GPUs with 80GB of memory.

\section{\method Benchmark Construction}
\label{app:clegr_construction}

\subsection{\method}
\subsubsection{Overview} 
The \method dataset is a graph-based question answering benchmark built on synthetic subway networks extending on the work by ~\cite{mack2018clevrgraph}. Each graph represents a fictional subway system with randomly generated nodes and attributes. Questions are categorized into two types: \textit{Fact-Based} and \textit{Reasoning}.

\subsubsection{Graph Generation}
Graphs in \method are primarily constructed using \textit{lines}. Each graph begins with a predefined number of lines, where each line is assigned a unique name and the following attributes:
\begin{itemize}
    \item \texttt{has\_aircon}
    \item \texttt{color}
    \item \texttt{stroke}
    \item \texttt{built}
\end{itemize}

Lines may intersect with each other. Along each line, a sequence of \textit{stations} is generated. Stations near line intersections may belong to two lines. Each station is assigned a unique name and the following attributes:
\begin{itemize}
    \item \texttt{disabled\_access}
    \item \texttt{has\_rail}
    \item \texttt{music}
    \item \texttt{architecture}
    \item \texttt{size}
    \item \texttt{cleanliness}
\end{itemize}

Edges connect every pair of adjacent stations on a line, and inherit properties from the parent line. The overall size of the graph is controlled by the number of lines and the number of stations per line.

\subsection{Graph Statistics} 
\label{app:graph_stats}

Table~\ref{tab:graph_stats} mentions the graph statistics for \method Subway Networks.

\begin{table}[H]
  \centering
  \caption{Graph statistics for \method Subway Networks with different graph sizes.}
  \small
  \setlength{\tabcolsep}{8pt}
  \begin{tabular}{lcc}
    \toprule
    \textbf{Metric} & \textbf{\method} & \textbf{\method-Large} \\
    \midrule
    Average Number of Nodes & 26.54 $\pm$ 5.41 & 73.17 $\pm$ 13.55 \\
    Average Number of Edges & 28.32 $\pm$ 6.37 & 78.87 $\pm$ 16.41 \\
    Average Number of Lines & 6.02 $\pm$ 1.23 & 9.97 $\pm$ 2.02 \\
    \bottomrule
  \end{tabular}

  \label{tab:graph_stats}
\end{table}

\subsection{Dataset Generation Pipeline}

\subsubsection{Questions}
The dataset contains 22 question templates for Fact Based questions and 34 question templates for Reasoning Based questions. Each generated graph will have two instances of each question template. The question templates are of the following form:
\begin{center}
    \textit{"How many stations playing \{\} does \{\} pass through?"}
\end{center}
The empty \{\} in the above template will be filled by a randomly picked Music and Line.

\subsubsection{Fact Based Questions}
Contains 22 questions which require retrieving information about nodes, edges or the graph itself. (All the templates can be viewed in Table~\ref{tab:fact_templates})

\subsubsection{Reasoning Based Questions}
Contains 34 questions which require the model to do some amount of reasoning on the information it retrieves from the graph. The different types of reasoning subgroups we try to incorporate in our dataset are listed here: \textbf{Aggregation}, \textbf{Filtering}, \textbf{PathReasoning}, \textbf{Topology}. (All the templates as per the question scope can be viewed at Table~\ref{tab:node_question_templates}, Table~\ref{tab:edge_question_templates}, and Table~\ref{tab:subgraph_question_templates}.)

\section{Dataset Generation and Schema}
\label{sec:appendix_dataset}

The dataset consists of procedurally generated transit (metro system) graphs, each accompanied by a set of questions and answers derived from its structure and attributes.

\subsection{Graph Generation Pipeline}
\label{sec:appendix_graph_gen}
Each graph, representing a unique transit map, is generated through a multi-stage pipeline designed to create complex and semi-realistic structures. The core generation logic is implemented in \texttt{generate\_graph.py}.

\begin{enumerate}
    \item \textbf{Line Generation:} A set of metro lines is created. Each line is assigned a unique ID, a name (e.g., "Blue Line", "Circle Express"), and a set of properties such as color, stroke style (solid, dashed, dotted), year of construction, and whether it has air conditioning. To ensure visual distinctiveness, combinations of color and stroke style are unique across the graph.

    \item \textbf{Station Placement along Curves:} For each line, a cubic B\'ezier curve is generated with random control points within a predefined map radius. A specified number of initial station locations are then calculated by evaluating points along this curve. This method produces smooth, winding paths for metro lines rather than simple straight lines. A small amount of Gaussian noise is added to each station's coordinates for organic variation. Each station is initialized with a unique, programmatically generated name and a set of properties (e.g., architecture, cleanliness, disabled access).

    \item \textbf{Station Coalescing:} A critical step to create realistic interchanges. A KD-Tree is used to efficiently find all stations across all lines that are within a minimum distance threshold of each other. These nearby stations are merged into a single node using a Disjoint Set Union (DSU) algorithm. The resulting merged node, representing an interchange station, inherits the ID and properties of one of its constituent pre-merge stations. This process transforms a simple collection of lines into a more complex, interconnected network.

    \item \textbf{Edge Generation:} After stations are coalesced, edges are created to connect consecutive stations along each line's path. If a sequence of stations on a line was A → B → C and stations B and C were coalesced into a new station D, the resulting edges would connect A → D. Edges store the IDs of the two stations they connect and inherit properties from their parent line, such as its name, color, and stroke style.

    \item \textbf{Connectivity Assurance:} The graph is checked for connectivity using NetworkX. If the graph consists of multiple disconnected components, new "connector" edges are added to link them. A random node is chosen from each of two components, and a new edge is created between them. This edge is styled as a dotted line to be visually distinct and is assigned to one of the existing lines. This ensures the entire graph is a single connected component, which is a prerequisite for many graph algorithms (e.g., shortest path calculations between any two nodes).
    
    \item \textbf{Integer Naming (Optional):} For certain model training regimes, all human-readable station and line names can be replaced with unique integer strings. This prevents models from learning spurious correlations from the names themselves and forces them to rely solely on the graph's topology and categorical features. When this option is enabled, an entity's ID is also updated to match its new integer name for consistency.
\end{enumerate}

The entire generation process is configurable via command-line arguments, allowing for the creation of graphs of varying sizes (small, medium, large, or a random mix), controlled by parameters like the number of lines, stations per line, and the map radius.

\subsection{Feature Schema}
The dataset contains three primary entities: Nodes (Stations), Edges (Tracks), and Lines. Their attributes are detailed below.

\subsubsection{Node Features (Stations)}
Each node represents a station and has the following attributes, which are one-hot encoded to form the node feature tensor \texttt{x}.
\begin{description}
    \item[\texttt{id}] (String): A unique identifier for the station. If integer names are used, this is the integer as a string.
    \item[\texttt{name}] (String): The human-readable or integer name of the station.
    \item[\texttt{x, y}] (Float): The 2D coordinates of the station on the map.
    \item[\texttt{disabled\_access}] (Boolean): Whether the station has disabled access.
    \item[\texttt{has\_rail}] (Boolean): A categorical property, e.g., for distinguishing train stations from bus stations in a mixed-modal system.
    \item[\texttt{music}] (String): The genre of ambient music played (e.g., \texttt{'classical'}, \texttt{'rock'}, \texttt{'none'}).
    \item[\texttt{architecture}] (String): The architectural style of the station (e.g., \texttt{'victorian'}, \texttt{'modernist'}).
    \item[\texttt{size}] (String): The relative size of the station (e.g., \texttt{'small'}, \texttt{'large'}).
    \item[\texttt{cleanliness}] (String): A binary property (\texttt{'clean'} or \texttt{'dirty'}).
\end{description}

\subsubsection{Edge Features (Tracks)}
Each edge represents a track segment between two stations on a specific line. Edges are directed in the PyG representation (i.e., an edge from A to B is distinct from B to A), but represent an undirected physical connection. Their attributes form the edge feature tensor \texttt{edge\_attr}.
\begin{description}
    \item[\texttt{station1, station2}] (String): The IDs of the nodes connected by the edge.
    \item[\texttt{line\_id}] (String): The ID of the line this track segment belongs to.
    \item[\texttt{line\_name}] (String): The name of the line.
    \item[\texttt{line\_color}] (String): The color of the line.
    \item[\texttt{line\_stroke}] (String): The stroke style of the line (e.g., \texttt{'solid'}, \texttt{'dotted'}).
    \item[\texttt{properties}] (Dict): A dictionary containing properties inherited from the line, such as \texttt{'line\_has\_aircon'} (Boolean) and \texttt{'line\_built'} (String, e.g., '1990').
\end{description}

\subsection{Question-Answer Generation}
With a graph fully generated, the \texttt{generate.py} script produces a set of question-answer pairs.
\begin{enumerate}
    \item \textbf{Question Templates:} A predefined set of \texttt{QuestionForm} objects encapsulate different types of questions. These are organized by group (e.g., lookup, comparison) and type (e.g., existence, counting).
    \item \textbf{Instantiation:} For each graph, the script iterates a specified number of times to generate questions. In each iteration, it randomly selects a \texttt{QuestionForm} and attempts to instantiate it using the current graph. This involves sampling nodes, lines, or properties from the graph to fill in the template's parameters.
    \item \textbf{Answer Derivation:} The ground truth answer is derived by programmatically executing a functional representation of the question on the \texttt{GraphSpec} object. For example, to answer "How many stations on the Red Line are large?", the program iterates through the nodes on the Red Line and counts how many have their \texttt{size} attribute set to \texttt{'large'}. If a question cannot be instantiated (e.g., a question about interchanges in a graph with none), the attempt is discarded, and another form is tried.
\end{enumerate}

This process yields a diverse set of questions, ranging from simple property lookups ("Does Red Station have disabled access?") to complex multi-hop reasoning involving counting, comparison, and logical operations ("Which line has more modernist stations, the Blue Line or the Green Line?").

\subsection{Final Data Format}
The complete dataset is saved as a list of PyTorch Geometric \texttt{Data} objects in a single \texttt{.pt} file. Crucially, \textbf{each \texttt{Data} object represents a single graph-question-answer triplet}.

A companion \texttt{\_mappers.pkl} file is also saved, containing dictionaries that map the raw string values of all categorical features to their integer indices used in the feature tensors.

Each \texttt{torch\_geometric.data.Data} object has the following key attributes:
\begin{description}
    \item[\texttt{x}] (Tensor): Node feature matrix of shape $[N, F_{node}]$, where $N$ is the number of nodes and $F_{node}$ is the size of the embedding of the sentence representing the node features encoded by BERT.
    \item[\texttt{edge\_index}] (Tensor): Graph connectivity in COO format, a tensor of shape $[2, E]$, where $E$ is the number of directed edges.
    \item[\texttt{edge\_attr}] (Tensor): Edge feature matrix of shape $[E, F_{edge}]$, where $F_{edge}$ is the size of the embedding of the sentence representing the edge features encoded by BERT.
    \item[\texttt{question}] (String): The natural language question, e.g., "How many stations are on the Cyan Line?".
    \item[\texttt{label}] (String): The ground truth answer, serialized to a string (e.g., \texttt{'12'}, \texttt{'True'}, \texttt{'Red Line'}).
    \item[\texttt{question\_type}] (String): A unique string identifying the question template used, e.g., \texttt{'CountStationsOnLine'}.
    \item[\texttt{question\_group}] (String): The general category of the question, e.g., \texttt{'count'}.
\end{description}

\subsubsection{Node Sentence Representation}
The textual attributes of nodes are used to generate a sentence of the following form describing the node:

\textit{\{NodeName\} \{has/does not have\} disabled access and \{has/does not have\} rail. It features \{Architecture\} architecture, has \{Cleanliness\} cleanliness, \{Music\} music and is \{Size\} in size.}

\subsubsection{Edge Sentence Representation}
Similarly, the textual attributes of edges are used to generate a sentence of the following form:

\textit{There is a \{LineStroke\} \{LineColour\} line from \{SourceStation\} to \{DestinationStation\}. It \{has/does not have\} air conditioning and was built in \{BuiltYear\}.}

\subsubsection{Graph Embeddings}
The sentence representation of each node/edge is encoded into a 768 dimensional embedding using \texttt{bert-base-uncased}. These embeddings are passed to the GNN backbone.

\subsubsection{Data Tuples}
Each Data example in \method is a tuple of the form (Graph, Question, Answer). 

\subsubsection{Dataset Statistics}
The total default number of graphs generated is 500. The Small \method dataset contains 22000 Fact Based questions(44 Fact Based questions per graph)  and 32248 (after natural filteration of invalid questions from a total of 34000 questions)  (To check and update) Reasoning questions(68 Reasoning Based questions per graph). The Train, Validation and Test set contain 300, 100 and 100 of the graphs(and their corresponding questions) respectively.

\subsubsection{Output Format}
The model's output is of one of the following formats: List, Boolean, String, Numeric. The prompt passed to the model will be suffixed with text describing the output format.

\begin{table}[t]
\caption{\method\ Fact-Based Question Template Definitions for the Subway Networks}
\label{tab:fact_templates}
\centering
\small
\renewcommand{\arraystretch}{1.15}
\setlength{\tabcolsep}{3pt} %
\begin{tabularx}{\linewidth}{@{}%
>{\raggedright\arraybackslash}p{3.6cm}  %
>{\raggedright\arraybackslash}X          %
>{\centering\arraybackslash}p{1.8cm}     %
>{\centering\arraybackslash}p{1.4cm}     %
@{}}
\toprule
\textbf{Template Name} & \textbf{Question Template} & \textbf{Output Type} & \textbf{Scope} \\
\midrule
StationPropertyCleanliness   & How clean is \textcolor{myblue}{\{Station\}}? & String  & Node \\
StationPropertyCleanliness2  & What is the cleanliness level of \textcolor{myblue}{\{Station\}} station? & String  & Node \\
StationPropertySize          & How big is \textcolor{myblue}{\{Station\}}? & String  & Node \\
StationPropertySize2         & What size is \textcolor{myblue}{\{Station\}}? & String  & Node \\
StationPropertyMusic         & What music plays at \textcolor{myblue}{\{Station\}}? & String  & Node \\
StationPropertyMusic2        & Which type of music is played at \textcolor{myblue}{\{Station\}}? & String  & Node \\
StationPropertyArchitecture  & What architectural style is \textcolor{myblue}{\{Station\}}? & String  & Node \\
StationPropertyArchitecture2 & Describe \textcolor{myblue}{\{Station\}} station's architectural style. & String  & Node \\
StationPropertyDisabled
\\Access & Does \textcolor{myblue}{\{Station\}} have disabled access? & Boolean & Node \\
StationPropertyDisabled
\\Access2 & Is there disabled access at \textcolor{myblue}{\{Station\}}? & Boolean & Node \\
StationPropertyHasRail       & Does \textcolor{myblue}{\{Station\}} have rail connections? & Boolean & Node \\
StationPropertyHasRail2      & Can you get rail connections at \textcolor{myblue}{\{Station\}}? & Boolean & Node \\
StationExistence1            & Is there a station called \textcolor{myblue}{\{Station\}}? & Boolean & Node \\
StationExistence2            & Is there a station called \textcolor{myblue}{\{FakeStationName\}}? & Boolean & Node \\
StationLine                  & Which lines is \textcolor{myblue}{\{Station\}} on? & List    & Edge \\
StationLineCount             & How many lines is \textcolor{myblue}{\{Station\}} on? & Numeric & Edge \\
StationAdjacentAlwaysTrue    & Are \textcolor{myblue}{\{Station\}} and \textcolor{myblue}{\{Station\}} adjacent? & Boolean & Edge \\
StationAdjacent              & Are \textcolor{myblue}{\{Station\}} and \textcolor{myblue}{\{Station\}} adjacent? & Boolean & Edge \\
EdgePropertyColor            & What color is the line between \textcolor{myblue}{\{Station\}} and \textcolor{myblue}{\{Station\}}? & String  & Edge \\
EdgePropertyAircon           & Does the line between \textcolor{myblue}{\{Station\}} and \textcolor{myblue}{\{Station\}} have air conditioning? & Boolean & Edge \\
EdgePropertyStroke           & What stroke style is the line between \textcolor{myblue}{\{Station\}} and \textcolor{myblue}{\{Station\}}? & String  & Edge \\
EdgePropertyBuilt            & When was the line between \textcolor{myblue}{\{Station\}} and \textcolor{myblue}{\{Station\}} built? & String  & Edge \\
\bottomrule
\end{tabularx}
\end{table}

\begin{table}[t]
\caption{\method\ Reasoning-Based Question Templates for the Subway Networks (Scope: Node)}
\label{tab:node_question_templates}
\centering
\small
\renewcommand{\arraystretch}{1.15}
\setlength{\tabcolsep}{3pt} %
\begin{tabularx}{\linewidth}{@{}%
>{\raggedright\arraybackslash}p{3.9cm}  %
>{\raggedright\arraybackslash}X          %
>{\centering\arraybackslash}p{1.9cm}     %
@{}}
\toprule
\textbf{Template Name} & \textbf{Question Template} & \textbf{Output Type} \\
\midrule
StationPairAdjacent        & Which station is adjacent to both \textcolor{myblue}{\{Station\}} and \textcolor{myblue}{\{Station\}}? & String \\
StationArchitectureAdjacent& Which \textcolor{myblue}{\{Architecture\}} station is adjacent to \textcolor{myblue}{\{Station\}}? & String \\
StationTwoHops             & How many other stations are two stops or closer to \textcolor{myblue}{\{Station\}}? & Numeric \\
HasCycle                   & Is \textcolor{myblue}{\{Station\}} part of a cycle? & Boolean \\
StationOneApartTrue        & Are \textcolor{myblue}{\{Station\}} and \textcolor{myblue}{\{Station\}} connected by the same station? & Boolean \\
StationOneApart            & Are \textcolor{myblue}{\{Station\}} and \textcolor{myblue}{\{Station\}} connected by the same station? & Boolean \\
TopologyMostCommonArch     & What is the most common architectural style of stations within 2 hops of \textcolor{myblue}{\{Station\}}? & String \\
CountIntersectionProperties& How many stations are both large and have disabled access? & Numeric \\
CompareArchitectureCount   & Which architectural style has more stations, \textcolor{myblue}{\{Architecture\}} or \textcolor{myblue}{\{Architecture\}}? & String \\
\bottomrule
\end{tabularx}
\end{table}

\begin{table}[t]
\caption{\method\ Reasoning-Based Question Templates for the Subway Networks (Scope: Edge)}
\label{tab:edge_question_templates}
\centering
\small
\renewcommand{\arraystretch}{1.15}
\setlength{\tabcolsep}{3pt} %
\begin{tabularx}{\linewidth}{@{}%
>{\raggedright\arraybackslash}p{3.9cm}  %
>{\raggedright\arraybackslash}X          %
>{\centering\arraybackslash}p{1.9cm}     %
@{}}
\toprule
\textbf{Template Name} & \textbf{Question Template} & \textbf{Output Type} \\
\midrule
StationSameLineTrue   & Are \textcolor{myblue}{\{Station\}} and \textcolor{myblue}{\{Station\}} on the same line? & Boolean \\
EdgeFilterAirconCount & How many air-conditioned lines is \textcolor{myblue}{\{Station\}} connected to? & Numeric \\
EdgeFilterColorCount  & How many \textcolor{myblue}{\{Color\}} lines is \textcolor{myblue}{\{Station\}} connected to? & Numeric \\
PathYearSpan          & How many years newer is the newest line between \textcolor{myblue}{\{Station\}} and \textcolor{myblue}{\{Station\}} compared to the oldest? & Numeric \\
PathOptimalColor      & What is the most common line color on the shortest path between \textcolor{myblue}{\{Station\}} and \textcolor{myblue}{\{Station\}}? & String \\
PathEarliestBuilt     & What is the earliest year a line was built on the shortest path between \textcolor{myblue}{\{Station\}} and \textcolor{myblue}{\{Station\}}? & String \\
\bottomrule
\end{tabularx}
\end{table}

\begin{table}[t]
\caption{\method\ Reasoning-Based Question Templates for the Subway Networks (Scope: Sub-graph)}
\label{tab:subgraph_question_templates}
\centering
\small
\renewcommand{\arraystretch}{1.15}
\setlength{\tabcolsep}{3pt} %
\begin{tabularx}{\linewidth}{@{}%
>{\raggedright\arraybackslash}p{3.9cm}  %
>{\raggedright\arraybackslash}X          %
>{\centering\arraybackslash}p{1.9cm}     %
@{}}
\toprule
\textbf{Template Name} & \textbf{Question Template} & \textbf{Output Type} \\
\midrule
LineTotalArchitectureCount       & How many architectural styles does \textcolor{myblue}{\{Line\}} pass through? & Numeric \\
LineTotalMusicCount               & How many music styles does \textcolor{myblue}{\{Line\}} pass through? & Numeric \\
LineTotalSizeCount                & How many sizes of station does \textcolor{myblue}{\{Line\}} pass through? & Numeric \\
LineFilterMusicCount              & How many stations playing \textcolor{myblue}{\{Music\}} does \textcolor{myblue}{\{Line\}} pass through? & Numeric \\
LineFilterCleanlinessCount        & How many \textcolor{myblue}{\{Cleanliness\}} stations does \textcolor{myblue}{\{Line\}} pass through? & Numeric \\
LineFilterSizeCount               & How many \textcolor{myblue}{\{Size\}} stations does \textcolor{myblue}{\{Line\}} pass through? & Numeric \\
LineFilterDisabledAccessCount     & How many stations with disabled access does \textcolor{myblue}{\{Line\}} pass through? & Numeric \\
LineFilterHasRailCount            & How many stations with rail connections does \textcolor{myblue}{\{Line\}} pass through? & Numeric \\
LineStations                      & Which stations does \textcolor{myblue}{\{Line\}} pass through? & List \\
StationShortestCount              & How many stations are between \textcolor{myblue}{\{Station\}} and \textcolor{myblue}{\{Station\}}? & Numeric \\
StationShortestAvoidingCount      & How many stations are on the shortest path between \textcolor{myblue}{\{Station\}} and \textcolor{myblue}{\{Station\}} avoiding \textcolor{myblue}{\{Cleanliness\}} stations? & Numeric \\
StationShortestAvoiding
\\ArchitectureCount & How many stations are on the shortest path between \textcolor{myblue}{\{Station\}} and \textcolor{myblue}{\{Station\}} avoiding \textcolor{myblue}{\{Architecture\}} architecture stations? & Numeric \\
DistinctRoutes                    & How many distinct routes are there between \textcolor{myblue}{\{Station\}} and \textcolor{myblue}{\{Station\}}? & Numeric \\
CountEqualSizeStation             & How many stations in \textcolor{myblue}{\{Line\}} are of the same size as \textcolor{myblue}{\{Station\}}? & Numeric \\
LineIntersectionStations          & How many stations are shared between the \textcolor{myblue}{\{Line\}} and the \textcolor{myblue}{\{Line\}}? & Numeric \\
NodeOnPath                        & Is \textcolor{myblue}{\{Station\}} on the shortest path between \textcolor{myblue}{\{Station\}} and \textcolor{myblue}{\{Station\}}? & Boolean \\
PathMostCommonMusic               & What is the most common music style on the shortest path between \textcolor{myblue}{\{Station\}} and \textcolor{myblue}{\{Station\}}? & String \\
CompareLineDisabledAccess         & Which line has more stations with disabled access, \textcolor{myblue}{\{Line\}} or \textcolor{myblue}{\{Line\}}? & String \\
\bottomrule
\end{tabularx}
\end{table}

\subsection{\method Computer-Networks}
\label{app:computer_networks}

The \method Computer-Networks is a dataset containing graphs representing fictional computer networks. These graphs are generated very similar to the generation used in the Subway Networks dataset. There are some differences in the graph generation primarily due to the fact that "lines" do not exist in this dataset. The differences are highlighted below:

\begin{enumerate}
    \item \textbf{Node-Centric Foundation (vs. Line-Centric):} The process begins by generating a target number of nodes, as specified by the \texttt{nodes} parameter. Unlike the transit map's structured placement along B\'ezier curves, these system nodes are scattered with \textbf{random (x, y) coordinates} across the map space. The concept of a `LineSpec` is entirely absent; nodes are the primary, independent entities from the outset.

    \item \textbf{Emergent Topology via Proximity (vs. Pre-Defined Paths):} Edge creation is not determined by a sequential path. Instead, the \texttt{gen\_edges} function implements a k-nearest neighbor algorithm. After the final node positions are set, a \texttt{KDTree} is used to find the `k` closest neighbors for each node (where `k` is derived from the \texttt{avg\_degree} parameter). Edges are then created between a node and its neighbors.

    \item \textbf{Hub Formation (vs. Interchange Creation):} The function \texttt{coalesce\_nearby\_nodes} serves a different conceptual purpose here. In the transit model, it created interchanges by merging nodes from different lines. Here, it addresses the potential for random placement to create unrealistic clusters of nodes. By merging nodes that are too close, such node clusters are reduced.

    \item \textbf{Intrinsic Edge Properties (vs. Inherited):} In the transit model, edge properties (like color and stroke) were inherited from their parent line. In this dataset, every edge is assigned its own set of properties directly and randomly from the \texttt{EdgeProperties} dictionary. Attributes like \texttt{bandwidth\_units}, \texttt{latency\_ms}, and \texttt{encryption\_status} are intrinsic to the connection itself.
\end{enumerate}

\subsection{Feature Schema}
The feature schema is completely redesigned to reflect the computer network domain.

\subsubsection{Node Features (System Nodes)}
Each node represents a computer system, server, or device. Its attributes are one-hot encoded into the node feature tensor \texttt{x}.
\begin{description}
    \item[\texttt{id}] (String): A unique identifier for the system node.
    \item[\texttt{name}] (String): The programmatically generated name of the node.
    \item[\texttt{x, y}] (Float): The 2D coordinates of the node in the grid space.
    \item[\texttt{status}] (String): The operational status of the node (e.g., \texttt{'Operational'}, \texttt{'Offline'}, \texttt{'Overloaded'}).
    \item[\texttt{security\_level}] (String): An assigned security clearance level (e.g., \texttt{'Public'}, \texttt{'Internal'}, \texttt{'Restricted'}).
    \item[\texttt{location\_sector}] (String): The logical or physical sector where the node is located (e.g., \texttt{'Sector\_Red'}).
    \item[\texttt{firmware\_version}] (String): The version of the node's firmware (e.g., \texttt{'v1.1'}, \texttt{'v2.0'}).
    \item[\texttt{power\_consumption\_units}] (Integer): A measure of the node's power draw.
\end{description}

\subsubsection{Edge Features (Connections)}
Each edge represents a network connection between two system nodes. There is no `LineSpec` entity. Edge attributes are one-hot encoded into the edge feature tensor \texttt{edge\_attr}.
\begin{description}
    \item[\texttt{node1\_id, node2\_id}] (String): The IDs of the two nodes being connected. The class attribute is named station1, station2 to be compatible with the other domain's codebase.
    \item[\texttt{bandwidth\_units}] (Integer): A measure of the connection's data throughput capacity.
    \item[\texttt{latency\_ms}] (Integer): The latency of the connection in milliseconds.
    \item[\texttt{encryption\_status}] (String): The encryption state of the connection (e.g., \texttt{'Encrypted'}, \texttt{'Unencrypted'}).
\end{description}

\subsubsection{Node Sentence Representation}
Similar to the Subway Netowrks dataset, the textual attributes of nodes are used to generate a sentence of the following form describing the node:

\textit{System node \{NodeName\} is in \{LocationSector\} with status \{Status\}. It has security level \{SecurityLevel\}, firmware \{FirmwareVersion\}, and consumes \{PowerConsumptionUnits\} power units.}

\subsubsection{Edge Sentence Representation}
Similarly, the textual attributes of edges are used to generate a sentence of the following form:

\textit{A link connects \{SourceNodeName\} and \{DestinationNodeName\}. It has \{BandwidthUnits\} bandwidth units, \{Latency\}ms latency, and its encryption status is \{EncryptionStatus\}.}

\subsection{Graph Statistics}
\label{app:comp_nets}

Table~\ref{tab:graph_stats_comp} shows the graph statistics for \method Computer Networks, which is also similar to that of \method Subway Networks.

\begin{table}[H]
  \centering
  \caption{Graph statistics for \method Computer Networks with different graph sizes.}
  \small
  \setlength{\tabcolsep}{8pt}
  \begin{tabular}{lcc}
    \toprule
    \textbf{Metric} & \textbf{\method Computer-Networks} \\
    \midrule
    Average Number of Nodes & 21.64 ± 3.93 \\
    Average Number of Edges & 28.61 ± 5.30 \\
    \bottomrule
  \end{tabular}

  \label{tab:graph_stats_comp}
\end{table}

All the question templates for the \method Computer Networks are present at Table~\ref{tab:fact_templates_grid} \& Table~\ref{tab:reasoning_templates_grid}

\begin{table}[t]
\caption{Fact-Based Question Template Definitions for the \method\ Computer-Networks Dataset}
\label{tab:fact_templates_grid}
\centering
\small
\renewcommand{\arraystretch}{1.15}
\setlength{\tabcolsep}{3pt} %
\begin{tabularx}{\linewidth}{@{}%
>{\raggedright\arraybackslash}p{3.6cm}  %
>{\raggedright\arraybackslash}X          %
>{\centering\arraybackslash}p{1.9cm}     %
>{\centering\arraybackslash}p{1.6cm}     %
@{}}
\toprule
\textbf{Template Name} & \textbf{Question Template} & \textbf{Output Type} & \textbf{Scope} \\
\midrule
NodePropertyStatus       & What is the status of node \textcolor{myblue}{\{Node\}}? & String  & Node \\
NodePropertySecurity     & What is the security level of node \textcolor{myblue}{\{Node\}}? & String  & Node \\
NodePropertyLocation     & Which sector is node \textcolor{myblue}{\{Node\}} located in? & String  & Node \\
NodePropertyFirmware     & What firmware version runs on \textcolor{myblue}{\{Node\}}? & String  & Node \\
NodePropertyPower        & How many power units does \textcolor{myblue}{\{Node\}} consume? & Numeric & Node \\
NodeExistence1           & Is there a node named \textcolor{myblue}{\{Node\}} in the grid? & Boolean & Node \\
NodeExistence2           & Is there a node named \textcolor{myblue}{\{FakeNodeName\}} in the grid? & Boolean & Node \\
NodeAdjacentTrue         & Are nodes \textcolor{myblue}{\{Node\}} and \textcolor{myblue}{\{Node\}} directly linked? & Boolean & Edge \\
NodeAdjacent             & Are nodes \textcolor{myblue}{\{Node\}} and \textcolor{myblue}{\{Node\}} directly linked? & Boolean & Edge \\
\bottomrule
\end{tabularx}
\end{table}

\begin{table}[t]
\caption{Reasoning-Based Question Template Definitions for the \method\ Computer-Networks Dataset}
\label{tab:reasoning_templates_grid}
\centering
\small
\renewcommand{\arraystretch}{1.15}
\setlength{\tabcolsep}{3pt}
\begin{tabularx}{\linewidth}{@{}%
>{\raggedright\arraybackslash}p{3.6cm}
>{\raggedright\arraybackslash}X
>{\centering\arraybackslash}p{1.9cm}
>{\centering\arraybackslash}p{1.6cm}
@{}}
\toprule
\textbf{Template Name} & \textbf{Question Template} & \textbf{Output Type} & \textbf{Scope} \\
\midrule
CountNodesWithStatus        & How many nodes have status \textcolor{myblue}{\{Status\}}? & Numeric & Sub-graph \\
ListNodesInSector           & List all nodes in \textcolor{myblue}{\{Sector\}}. & List & --- \\
MostCommonFirmware          & What is the most common firmware version? & String & Sub-graph \\
CountNodesWithTwoProps      & How many nodes in \textcolor{myblue}{\{Sector\}} have security level \textcolor{myblue}{\{Security Level\}}? & Numeric & Sub-graph \\
CountNeighborsOperational   & How many neighbors of \textcolor{myblue}{\{Node\}} are 'Operational'? & Numeric & Sub-graph \\
ShortestPathLen              & How many nodes are on shortest path between \textcolor{myblue}{\{Node\}} and \textcolor{myblue}{\{Node\}}? & Numeric & Sub-graph \\
NodesBetween                 & How many nodes lie between \textcolor{myblue}{\{Node\}} and \textcolor{myblue}{\{Node\}} on that path? & Numeric & Sub-graph \\
PathAvoidingStatus           & Is there a path from \textcolor{myblue}{\{Node\}} to \textcolor{myblue}{\{Node\}} avoiding status \textcolor{myblue}{\{Status\}}? & Boolean & Sub-graph \\
WithinHops                   & How many other nodes are within 3 hops of \textcolor{myblue}{\{Node\}}? & Numeric & Node \\
HasCycle                     & Is \textcolor{myblue}{\{Node\}} part of a cycle? & Boolean & Node \\
OneIntermediary              & Are \textcolor{myblue}{\{Node\}} and \textcolor{myblue}{\{Node\}} connected via exactly one intermediary? & Boolean & Edge \\
\bottomrule
\end{tabularx}
\end{table}

\section{\method Evaluation}
\label{app:evaluation_details}

\subsection{Evaluation Metrics by Answer Type}

\method includes four distinct answer formats, each evaluated with tailored methods:

\begin{itemize}
  \item \textbf{Categorical Answers} (e.g., station names, architecture types):\\
  Evaluated using \textit{exact match} after normalization (lowercasing and punctuation removal).

  \item \textbf{Boolean Answers} (e.g., yes/no questions):\\
  Text responses such as ``yes'', ``no'', ``true'', or ``false'' are mapped to binary values. We compute \textit{accuracy}, \textit{F1-score}, and \textit{Matthews Correlation Coefficient (MCC)}.

  \item \textbf{Numeric Answers} (e.g., distances, years):\\
  Evaluated using approximate equality via \texttt{numpy.isclose}. If direct parsing fails, we extract numbers using regex. Metrics include \textit{accuracy}, \textit{Mean Absolute Error (MAE)}, and \textit{Root Mean Square Error (RMSE)}.

  \item \textbf{Set-Valued Answers} (e.g., lists of stations):\\
  Scored using set-based \textit{precision}, \textit{recall}, and \textit{F1-score}, based on overlap with the ground truth set.
\end{itemize}

\subsection{Overall Evaluation}

Each question is scored using the appropriate method for its answer type. We report \textbf{overall accuracy} as the primary evaluation metric, defined as the proportion of correctly answered questions across the full test set. This ensures a fair and consistent comparison across models and tasks.

\section{Prompt Templates}
\subsection{Example Prompts for NC, each dataset}
\label{app:prompts_nc}

\subsubsection{Arxiv} \
\begin{tcolorbox}[simplertakeaway, title=Textual Prompt Format]
    \textless
    s\textgreater
    [INST] Title: \{Title\}\textbackslash
    nAbstract: \{Abstract\}\textbackslash
    nAnswer the following question: Which subcategory does this paper belong to? Please only output the most likely answer from the following subcategories and nothing else: \{Comma seperated Category List\}. \textbackslash
    nAnswer: [/INST] \{Label\} [/s]
\end{tcolorbox}

\subsubsection{Cora} \
\begin{tcolorbox}[simplertakeaway, title=Textual Prompt Format]
    \textless
    s\textgreater
    [INST] Title: \{Title\}\textbackslash
    nAbstract: \{Abstract\}\textbackslash
    nAnswer the following question: Which subcategory does this paper belong to? Please only output the most likely answer from the following subcategories and nothing else: theory, reinforcement learning, genetic algorithms, neural networks, probabilistic methods, case based, rule learning. \textbackslash
    nAnswer: [/INST] \{Label\} [/s]
\end{tcolorbox}

\subsubsection{CiteSeer} \
\begin{tcolorbox}[simplertakeaway, title=Textual Prompt Format]
    \textless
    s\textgreater
    [INST] Text: \{Text\}\textbackslash
    nAnswer the following question: Which category does this paper belong to? Please only output the most likely answer from the following categories directly and nothing else: Agents, AI, DB, IR, ML, HCI. \textbackslash
    nAnswer: [/INST] \{Label\} [/s]
\end{tcolorbox}

\subsubsection{Computers} \
\begin{tcolorbox}[simplertakeaway, title=Textual Prompt Format]
    \textless
    s\textgreater
    [INST] \{Context\}\textbackslash
    nAnswer the following question: Which computer product subcategory does this review belong to? Please only output the most likely answer from the following subcategories and nothing else: computer accessories and peripherals, tablet accessories, laptop accessories, computers and tablets, computer components, data storage, networking products, monitors, servers, tablet replacement parts \textbackslash
    n\textbackslash
    n Answer: [/INST] \{Label\} [/s]
\end{tcolorbox}

\subsubsection{Photo} \
\begin{tcolorbox}[simplertakeaway, title=Textual Prompt Format]
    \textless
    s\textgreater
    [INST] \{Context\}\textbackslash
    nAnswer the following question: Which photography related subcategory does this description belong to? Please only output the most likely answer from the following subcategories and nothing else: Film Photography, Video, Digital Cameras, Accessories, Binoculars \& Scopes, Lenses, Bags \& Cases, Lighting \& Studio, Flashes, Tripods \& Monopods, Underwater Photography, Video Surveillance \textbackslash
    n\textbackslash
    n Answer: [/INST] \{Label\} [/s]
\end{tcolorbox}

\subsubsection{History} \
\begin{tcolorbox}[simplertakeaway, title=Textual Prompt Format]
    \textless
    s\textgreater
    [INST] \{Context\}\textbackslash
    nAnswer the following question: Which history related subcategory does this description belong to? Please only output the most likely answer from the following subcategories and nothing else: World, Americas, Asia, Military, Europe, Russia, Africa, Ancient Civilizations, Middle East, Historical Study \& Educational Resources, Australia \& Oceania, Arctic \& Antarctica \textbackslash
    n\textbackslash
    n Answer: [/INST] \{Label\} [/s]
\end{tcolorbox}

\subsubsection{\method Facts and \method Reasoning} \
\begin{tcolorbox}[simplertakeaway, title=Textual Prompt Format]
    \textless
    s\textgreater
    [INST] \verb|---| Nodes \verb|---|\textbackslash
    n\{CSV String describing all the Nodes\}\textbackslash
    n\verb|---| Edges \verb|---|\textbackslash
    n\{CSV String describing all the Edges\}Above is the representation of a synthetic subway network. All stations and lines are completely fictional. Keep in mind that the subway network is not real. All information necessary to answer the question is present in the above representation. The question is: \{Question\}\textbackslash
    n\textbackslash
    n\{Answer Format Suffix\} [/INST] \{Label\} [/s]
\end{tcolorbox}

\textbf{CSV String describing all the Nodes:} A string describing all the nodes formatted like a CSV files with rows representing each node and columns describing different attributes of the nodes following the G‑Retriever~\cite{he2024gretrieverretrievalaugmentedgenerationtextual} pipeline.
\begin{tcolorbox}[simplertakeaway, title=CSV Header]

    ``id", ``name", ``disabled\_access", ``has\_rail", ``architecture", ``cleanliness", ``music", ``size" \
    
\end{tcolorbox}

\textbf{CSV String describing all the Edges:} A string describing all the nodes formatted like a CSV files with rows representing each edge and columns describing different attributes of the edges.
\begin{tcolorbox}[simplertakeaway, title=CSV Header]

    ``source\_id", ``target\_id", ``line\_color", ``line\_stroke", ``has\_aircon", ``built" \
    
\end{tcolorbox}

\textbf{Answer Format Suffix:}
Depending on the question and the output format expected by the question, one of the following is suffixed to the question.
\begin{tcolorbox}[simplertakeaway, title=Output Format Suffixes]
    \textbf{String Output:} Answer directly: \\
    \textbf{Bool Output:} Answer with `True' or `False': \textbackslash
    n\textbackslash
    nAnswer:  \\
    \textbf{List Output:} Output a comma-separated list: \\
    \textbf{Count Output:} Answer with a number: \textbackslash
    n\textbackslash
    nAnswer:  \\
    \textbf{Cycle Detection Question:} Answer with `True' if it is in a cycle, otherwise `False': \textbackslash
    n\textbackslash
    nAnswer:  
    
\end{tcolorbox}

\section{Extended Experimental Results}
\label{app:extended_results}

\subsection{Node Classification (Zero-Shot Transfer)}

\begin{table}[t]
\caption{Zero-shot transfer accuracy (\%) from Computers to History and Photo.}
\label{tab:transfer_computers}
\centering
\small
\setlength{\tabcolsep}{6pt} %
\renewcommand{\arraystretch}{1.1}
\begin{tabular}{lcc}
\toprule
\textbf{Model} & Computers $\rightarrow$ History & Computers $\rightarrow$ Photo \\
\midrule
\multicolumn{3}{c}{\itshape Graph‐Language Models} \\
\tglm\ (Llama 8B)        & 57.94 ± 4.97  & 5.44 ± 2.50 \\
\tglm\ (Phi 3.5B)        & 18.16 ± 5.70  & 3.71 ± 0.16 \\
\grsagetok\ (Llama 8B)   & 25.65 ± 17.15 & 5.01 ± 1.65 \\
\grsagetok\ (Phi 3.5B)   & 19.62 ± 5.07  & 2.52 ± 0.22 \\
\midrule
\multicolumn{3}{c}{\itshape Soft‐Prompted LLMs} \\
\softllama               & 43.56 ± 12.50 & 6.00 ± 1.19 \\
\softphithree            & 32.85 ± 13.79 & 2.85 ± 0.42 \\
\midrule
\multicolumn{3}{c}{\itshape Random Baseline (1/num\_classes)} \\
Uniform Guessing         & 8.33          & 8.33 \\
\bottomrule
\end{tabular}
\end{table}

\begin{table}[t]
\caption{Zero-shot transfer accuracy (\%) from Arxiv to Cora.}
\label{tab:transfer_arxiv}
\centering
\small
\setlength{\tabcolsep}{8pt}
\renewcommand{\arraystretch}{1.1}
\begin{tabular}{lc}
\toprule
\textbf{Model} & Arxiv $\rightarrow$ Cora \\
\midrule
\multicolumn{2}{c}{\itshape Graph‐Language Models} \\
\tglm\ (Llama 8B)        & 10.02 ± 2.73 \\
\tglm\ (Phi 3.5B)        & 5.80 ± 4.73 \\
\grsagetok\ (Llama 8B)   & 5.12 ± 4.18 \\
\grsagetok\ (Phi 3.5B)   & 3.76 ± 4.32 \\
\midrule
\multicolumn{2}{c}{\itshape Soft‐Prompted LLMs} \\
\softllama               & 16.57 ± 1.15 \\
\softphithree            & 17.19 ± 0.68 \\
\midrule
\multicolumn{2}{c}{\itshape Random Baseline (1/num\_classes)} \\
Uniform Guessing         & 14.29 \\
\bottomrule
\end{tabular}
\end{table}

Effective zero-shot transfer learning is a key advantage claimed by GLMs like \tglm. To evaluate this, we perform zero-shot transfer experiments across distinct semantic domains (Computers to History and Photo, and Arxiv to Cora). Surprisingly, our results (Table~\ref{tab:transfer_computers} and Table~\ref{tab:transfer_arxiv}) indicate negligible gains for GLMs over soft-prompted LLM baselines, and in some cases, GLMs even perform worse. This finding challenges the claimed superior generalization of GLMs, highlighting that even in tasks considered advantageous for GLMs, simple soft-prompted LLMs exhibit comparable or superior performance, raising questions about the practical value of current GLM designs in zero-shot contexts of Node classification.

To achieve cross-dataset transfer capability, where models trained on one graph dataset can perform reasoning on different target datasets, we design instructions that include both the task description and the complete set of alternative answers from both source and target domains. This approach follows the methodology established by \tglm.

For the Arxiv → Cora transfer scenario, the training instruction is structured as: "Which research area does this paper belong to? Please directly give the most likely answer from the following categories: {ans}", where {ans} contains class labels from both the Arxiv dataset (computer science subcategories such as "cs.AI", "cs.LG") and the Cora dataset (machine learning areas such as "Neural Networks", "Reinforcement Learning").

Similarly, for the Computers → History and Computers → Photo transfers, the instruction follows: "Which product category does this item belong to? Please select from the following options: {ans}", where {ans} includes classes from the Computers dataset alongside the respective target dataset classes (History product types or Photo equipment categories).
Including alternative answers from both source and target datasets enables the model to learn the task of "selecting the correct answer from a given set according to the reasoning requirements" rather than memorizing dataset-specific label mappings, thus facilitating effective knowledge transfer across different graph domains.

\begin{table*}[htbp]
  \centering
  \caption{Zero-shot Transfer Accuracy: Trained on \method Subway Reasoning, Tested on \method Computer Network Reasoning}
  \normalsize
  \setlength{\tabcolsep}{10pt}
  \begin{tabular}{ccc}
    \toprule
    \textbf{Model} & \textbf{Trained $\rightarrow$ Tested} & \textbf{Accuracy} \\
    \midrule
    \sage  \llama & \method Subway $\rightarrow$ \method Computer & 33.16 ± 0.97 \\
    \sage  \phithree & \method Subway $\rightarrow$ \method Computer & 27.33 ± 2.51 \\
    \sage  \phifour & \method Subway $\rightarrow$ \method Computer & 39.3 ± 2.34 \\
    \tglm  \llama & \method Subway $\rightarrow$ \method Computer & 32.92 ± 2.27 \\
    \tglm  \phithree & \method Subway $\rightarrow$ \method Computer & 27.67 ± 2.52 \\
    \tglm  \phifour & \method Subway $\rightarrow$ \method Computer & 40.72 ± 2.87 \\
    \softllama & \method Subway $\rightarrow$ \method Computer & 30.89 ± 0.66 \\
    \softphithree & \method Subway $\rightarrow$ \method Computer & 26.81 ± 1.45 \\
    \softphifour & \method Subway $\rightarrow$ \method Computer & 36.79 ± 1.17 \\
    \bottomrule
  \end{tabular}

  \label{tab:clegr_transfer_subway}
\end{table*}

\begin{table*}[htbp]
  \centering
  \caption{In-Domain Accuracy: Trained and Tested on \method Computer Network Reasoning}
  \normalsize
  \setlength{\tabcolsep}{10pt}
  \begin{tabular}{ccc}
    \toprule
    \textbf{Model} & \textbf{Trained $\rightarrow$ Tested} & \textbf{Accuracy} \\
    \midrule
    \sage  \llama & \method Computer $\rightarrow$ \method Computer & 52.33 ± 1.34 \\
    \sage  \phithree & \method Computer $\rightarrow$ \method Computer & 47.53 ± 1.46 \\
    \tglm  \llama & \method Computer $\rightarrow$ \method Computer & 47.94 ± 4.32 \\
    \tglm  \phithree & \method Computer $\rightarrow$ \method Computer & 39.67 ± 4.39 \\
    \softllama & \method Computer $\rightarrow$ \method Computer & 48.22 ± 2.07 \\
    \softphithree & \method Computer $\rightarrow$ \method Computer & 46.38 ± 2.48 \\
    \bottomrule
  \end{tabular}
  \label{tab:clegr_indomain_computer}
\end{table*}

\begin{table}[htbp]
\caption{Train on \method\ Subway Facts, Test on \method\ Computer (Facts / Reasoning).}
\label{tab:subwayfacts_to_computer}
\centering
\small
\setlength{\tabcolsep}{4pt} %
\renewcommand{\arraystretch}{1.1}
\begin{tabularx}{\linewidth}{@{}lXc@{}}
\toprule
\textbf{Model (Method + LLM)} & \textbf{Train $\rightarrow$ Test} & \textbf{Accuracy} \\
\midrule
\sage\ \llama    & \method\ Subway Facts $\rightarrow$ \method\ Computer Facts     & 87.99 ± 1.55 \\
\sage\ \phithree & \method\ Subway Facts $\rightarrow$ \method\ Computer Facts     & 87.96 ± 1.63 \\
\sage\ \phifour  & \method\ Subway Facts $\rightarrow$ \method\ Computer Facts     & 99.49 ± 0.30 \\
\tglm\ \llama    & \method\ Subway Facts $\rightarrow$ \method\ Computer Facts     & 91.68 ± 1.67 \\
\tglm\ \phithree & \method\ Subway Facts $\rightarrow$ \method\ Computer Facts     & 82.65 ± 2.53 \\
\midrule
\sage\ \llama    & \method\ Subway Facts $\rightarrow$ \method\ Computer Reasoning & 34.34 ± 2.76 \\
\sage\ \phithree & \method\ Subway Facts $\rightarrow$ \method\ Computer Reasoning & 31.10 ± 0.90 \\
\sage\ \phifour  & \method\ Subway Facts $\rightarrow$ \method\ Computer Reasoning & 40.27 ± 2.24 \\
\tglm\ \llama    & \method\ Subway Facts $\rightarrow$ \method\ Computer Reasoning & 34.06 ± 2.09 \\
\tglm\ \phithree & \method\ Subway Facts $\rightarrow$ \method\ Computer Reasoning & 28.17 ± 2.98 \\
\bottomrule
\end{tabularx}
\end{table}

\begin{table}[htbp]
\caption{Train on \method\ Subway Reasoning, Test on \method\ Computer (Facts / Reasoning).}
\label{tab:subwayreasoning_to_computer}
\centering
\small
\setlength{\tabcolsep}{4pt}
\renewcommand{\arraystretch}{1.1}
\begin{tabularx}{\linewidth}{@{}lXc@{}}
\toprule
\textbf{Model (Method + LLM)} & \textbf{Train $\rightarrow$ Test} & \textbf{Accuracy} \\
\midrule
\sage\ \llama    & \method\ Subway Reasoning $\rightarrow$ \method\ Computer Facts     & 45.26 ± 35.78 \\
\sage\ \phithree & \method\ Subway Reasoning $\rightarrow$ \method\ Computer Facts     & 79.50 ± 5.98 \\
\sage\ \phifour  & \method\ Subway Reasoning $\rightarrow$ \method\ Computer Facts     & 90.53 ± 4.83 \\
\tglm\ \llama    & \method\ Subway Reasoning $\rightarrow$ \method\ Computer Facts     & 71.74 ± 5.81 \\
\tglm\ \phithree & \method\ Subway Reasoning $\rightarrow$ \method\ Computer Facts     & 76.71 ± 2.50 \\
\midrule
\sage\ \llama    & \method\ Subway Reasoning $\rightarrow$ \method\ Computer Reasoning & 26.06 ± 11.86 \\
\sage\ \phithree & \method\ Subway Reasoning $\rightarrow$ \method\ Computer Reasoning & 28.51 ± 2.45 \\
\sage\ \phifour  & \method\ Subway Reasoning $\rightarrow$ \method\ Computer Reasoning & 36.07 ± 2.12 \\
\tglm\ \llama    & \method\ Subway Reasoning $\rightarrow$ \method\ Computer Reasoning & 34.37 ± 0.94 \\
\tglm\ \phithree & \method\ Subway Reasoning $\rightarrow$ \method\ Computer Reasoning & 28.34 ± 1.40 \\
\bottomrule
\end{tabularx}
\end{table}

\begin{table}[htbp]
\caption{Cross-Modality Transfer Between \method\ Subway Facts and \method\ Subway Reasoning.}
\label{tab:subway_crossmodality}
\centering
\small
\setlength{\tabcolsep}{4pt}
\renewcommand{\arraystretch}{1.1}
\begin{tabularx}{\linewidth}{@{}lXc@{}}
\toprule
\textbf{Model (Method + LLM)} & \textbf{Train $\rightarrow$ Test} & \textbf{Accuracy} \\
\midrule
\sage\ \llama    & \method\ Subway Facts $\rightarrow$ \method\ Subway Reasoning & 35.66 ± 0.31 \\
\sage\ \phithree & \method\ Subway Facts $\rightarrow$ \method\ Subway Reasoning & 30.30 ± 2.33 \\
\sage\ \phifour  & \method\ Subway Facts $\rightarrow$ \method\ Subway Reasoning & 36.09 ± 2.13 \\
\tglm\ \llama    & \method\ Subway Facts $\rightarrow$ \method\ Subway Reasoning & 33.93 ± 0.76 \\
\tglm\ \phithree & \method\ Subway Facts $\rightarrow$ \method\ Subway Reasoning & 29.19 ± 0.12 \\
\midrule
\sage\ \llama    & \method\ Subway Reasoning $\rightarrow$ \method\ Subway Facts & 42.38 ± 33.81 \\
\sage\ \phithree & \method\ Subway Reasoning $\rightarrow$ \method\ Subway Facts & 62.86 ± 5.34 \\
\sage\ \phifour  & \method\ Subway Reasoning $\rightarrow$ \method\ Subway Facts & 72.28 ± 1.86 \\
\tglm\ \llama    & \method\ Subway Reasoning $\rightarrow$ \method\ Subway Facts & 69.52 ± 5.88 \\
\tglm\ \phithree & \method\ Subway Reasoning $\rightarrow$ \method\ Subway Facts & 52.86 ± 6.82 \\
\bottomrule
\end{tabularx}
\end{table}

\subsection{\method (Zero Shot Transfer)} 

This section presents the accuracy results of GLMs on different transfer learning scenarios across domains and knowledge types (Facts vs. Reasoning). We present 5 tables (Table~\ref{tab:clegr_transfer_subway}, Table~\ref{tab:clegr_indomain_computer}, Table~\ref{tab:subwayfacts_to_computer}, Table~\ref{tab:subwayreasoning_to_computer}, and Table~\ref{tab:subway_crossmodality}) testing various scenarios like Zero-Shot domain transfer (e.g. \method Subway Networks to \method Computer Networks) and Zero-Shot in-domain but question-type transfer (e.g. \method Facts to \method Reasoning and vice versa).

\end{document}